\newif\ifCONF\CONFfalse
\newif\ifARXIV\ARXIVfalse

\ARXIVtrue

\ifCONF
\documentclass{llncs}
\usepackage{amsmath,amsfonts,amssymb,algorithm,algorithmic,}
\fi
\ifARXIV
\documentclass[a4paper]{article}
\usepackage{amsmath,amsfonts,amssymb,amsthm,algorithm,algorithmic}
\fi

\newcommand{\citealp}[1]{\cite{#1}}

\newcommand{\citep}[1]{\cite{#1}}
\newcommand{\citet}[1]{\cite{#1}}
\newcommand{\citeyear}[1]{\cite{#1}}

\newcommand{\e}{\mathrm{e}}
\newcommand{\givn}{\mathrel{|}}

 \newcommand{\bX}{\mathbf{X}}		            
\newcommand{\R}{\mathbb{R}}			    
\newcommand{\PPP}{\mathcal{P}}			    
\newcommand{\BP}{\begin{proof}}			    
\ifCONF
\newcommand{\EP}{\qed\end{proof}}		    
\fi
\ifARXIV
\newcommand{\EP}{\end{proof}}
\fi
\newcommand{\dL}{\mathcal{L}}			    
\newcommand{\EXP}{\mathrm{E}}                       
\newcommand{\defin}[1]{\emph{#1}}                   
\newcommand{\hs}[1]{\ensuremath{\mathcal{#1}}}      
\newcommand{\vect}[1]{\ensuremath{\mathbf{#1}}}     
\newcommand{\mat}[1]{\ensuremath{\mathbf{#1}}}      
\DeclareMathOperator{\diag}{diag}                   

\makeatletter
\floatstyle{\ALG@floatstyle}
\newfloat{protocol}{htbp}{loa}
\floatname{protocol}{Protocol}

\newfloat{algorithm}{htbp}{loa}
\floatname{algorithm}{Algorithm}
\makeatother

\ifARXIV
\newtheorem{theorem}{Theorem}
\newtheorem*{theorem*}{Theorem}
\newtheorem{proposition}{Proposition}
\newtheorem{lemma}{Lemma}
\newtheorem{corollary}{Corollary}

\theoremstyle{definition}
\newtheorem{definition}{Definition}
\newtheorem*{remark}{Remark}
\fi

\ifCONF
\fi

\begin{document}
\title{Prediction with Expert Advice \\ under Discounted Loss}
\ifCONF
\author{Alexey Chernov and Fedor Zhdanov\thanks{
Fedor Zhdanov is a PhD student.
He proved Theorems~\ref{thm:linbound} and~\ref{thm:hilbound}
and also contributed to Theorem~\ref{thm:AAD}.
The authors are ordered alphabetically, not by contribution.
}
}
\institute{Computer Learning Research Centre and Department of Computer Science,\\
            Royal Holloway, University of London,
            Egham, Surrey, TW20 0EX, UK\\
\email{\{chernov,fedor\}@cs.rhul.ac.uk}
}
\fi
\ifARXIV
\author{Alexey Chernov and Fedor Zhdanov\\[1ex]
\normalsize  Computer Learning Research Centre,
\normalsize  Department of Computer Science\\
\normalsize  Royal Holloway, University of London,
\normalsize  Egham, Surrey TW20 0EX, UK\\
\normalsize\texttt{\{chernov,fedor\}@cs.rhul.ac.uk}
}
\fi

\maketitle
\thispagestyle{empty}

\begin{abstract}
We study prediction with expert advice
in the setting where the losses are accumulated with some
discounting and the impact of old losses can gradually vanish.
We generalize the Aggregating Algorithm and
the Aggregating Algorithm for Regression,
propose a new variant of exponentially weighted average algorithm,
and prove bounds on the cumulative discounted loss.
\end{abstract}

\section{Introduction}\label{introduction}

Prediction with expert advice is
a framework for online sequence prediction.
Predictions are made step by step.
The quality of each prediction (the discrepancy between
the prediction and the actual outcome) is evaluated
by a real number called loss.
The losses are accumulated over time.
In the standard framework
for prediction with expert advice
(see the monograph~\cite{Cesa-BianchiLugosi}
for a comprehensive review),
the losses from all steps are just summed.
In this paper, we consider a generalization
where older losses can be devalued;
in other words,
we use discounted cumulative loss.

Predictions are made by Experts and Learner
according to Protocol~\ref{prot:GenDisc}.
\begin{protocol}
  \caption{Prediction with expert advice under general discounting}
  \label{prot:GenDisc}
  \begin{algorithmic}
    \STATE $\dL_0:=0$.
    \STATE $\dL_0^\theta:=0$, $\theta \in \Theta$.
    \FOR{$t=1,2,\dots$}
      \STATE Accountant announces $\alpha_{t-1}\in(0,1]$.
      \STATE Experts announce $\gamma_t^\theta \in\Gamma$, $\theta \in \Theta$.
      \STATE Learner announces $\gamma_t\in\Gamma$.
      \STATE Reality announces $\omega_t\in\Omega$.
      \STATE $\dL_t^\theta:=\alpha_{t-1}\dL_{t-1}^\theta
                      +\lambda(\gamma_t^\theta,\omega_t)$, $\theta \in \Theta$.
      \STATE $\dL_t:=\alpha_{t-1}\dL_{t-1}+\lambda(\gamma_t,\omega_t)$.
    \ENDFOR
  \end{algorithmic}
\end{protocol}
In this protocol,
$\Omega$ is the set of possible outcomes and
$\omega_1,\omega_2,\omega_3\ldots$ is the sequence to predict;
$\Gamma$ is the set of admissible predictions,
and ${\lambda\colon\Gamma\times\Omega\to[0,\infty]}$
is the loss function.
The triple $(\Omega,\Gamma,\lambda)$ specifies
the game of prediction.
The most common examples are
the binary square loss, log loss, and absolute loss games.
They have $\Omega=\{0,1\}$ and $\Gamma=[0,1]$,
and their loss functions are
$\lambda^\mathrm{sq}(\gamma,\omega)=(\gamma-\omega)^2$,
$\lambda^\mathrm{log}(\gamma,0)=-\log (1-\gamma)$ and
$\lambda^\mathrm{log}(\gamma,1)=-\log \gamma$,
$\lambda^\mathrm{abs}(\gamma,\omega)=\lvert\gamma-\omega\rvert$,
respectively.

The players in the game of prediction
are Experts $\theta$ from some pool $\Theta$,
Learner, and also Accountant and Reality.
We are interested in (worst-case optimal) strategies for Learner,
and thus the game can be regarded as a two-player game,
where Learner opposes the other players.
The aim of Learner is to keep his total loss $\dL_t$
small as compared to the total losses $\dL_t^\theta$
of all experts $\theta\in\Theta$.

The standard protocol of prediction with expert advice
(as described in~\cite{VovkAS,Vovk:98})
is a special case of Protocol~\ref{prot:GenDisc}
where Accountant always announces $\alpha_t=1$, $t=0,1,2,\ldots$.
The new setting gives some more freedom
to Learner's opponents.

Another important special case is the exponential (geometric) discounting
$\alpha_t=\alpha\in(0,1)$.
Exponential discounting is widely used
in finance and economics (see, e.\,g.,~\citet{Muth1960}),
time series analysis (see, e.\,g.,~\citealp{Gardner2006}),
reinforcement learning~\cite{BartoSutton},
and other applications.
In the context of prediction with expert advice,
Freund and Hsu~\citet{Freund2008} noted that
the discounted loss provides an alternative to
``tracking the best expert'' framework~\citet{HerbsterTBE}.
Indeed, an exponentially discounted sum depends 
almost exclusively on the last $O(\log (1/\alpha))$ terms.
If the expert with the best one-step performance
changes at this rate,
then Learner observing the $\alpha$-dis\-count\-ed losses
will mostly follow predictions of the current best expert.
Under our more general discounting,
more subtle properties of best expert changes 
may be specified by varying the discount factor.
In particular, one can cause Learner to ``restart mildly'' 
giving $\alpha_t=1$ (or $\alpha_t\approx 1$) 
most of the time and $\alpha_t\ll 1$ at crucial moments.
\ifARXIV
(We prohibit $\alpha_t=0$ in the protocol,
since this is exactly the same as the stopping the current
game and starting a new, independent game;
on the other hand, the assumption $\alpha_t\ne 0$
simplifies some statements.)
\fi

Cesa-Bianchi and Lugosi~\cite[\S~2.11]{Cesa-BianchiLugosi}
discuss another kind of discounting
\begin{equation}\label{eq:CBLdiscount}
 L_T = \sum_{t=1}^T \beta_{T-t} l_t\,,
\end{equation}
where $l_t$ are one-step losses and $\beta_t$ are some decreasing
discount factors.
To see the difference, let us rewrite our definition in the same style:
\ifCONF
\begin{equation}\label{eq:beta-discount}
L_T=\alpha_{T-1}L_{T-1}+ l_T
=
\sum_{t=1}^T \alpha_{t}\cdots\alpha_{T-1}l_t
= \frac{1}{\beta_T}\sum_{t=1}^T \beta_{t}l_t\,,
\end{equation}
\fi
\ifARXIV
\begin{multline}\label{eq:beta-discount}
L_T=\alpha_{T-1}L_{T-1}+ l_T
=
\alpha_{T-2}\alpha_{T-1}L_{T-2}
+\alpha_{T-1}l_{T-1}
+l_T
= \ldots
\\
=
\sum_{t=1}^T \alpha_{t}\cdots\alpha_{T-1}l_t
= \frac{1}{\beta_T}\sum_{t=1}^T \beta_{t}l_t\,,
\end{multline}
\fi
where $\beta_t=1/\alpha_1\cdots\alpha_{t-1}$, $\beta_1=1$.
The sequence $\beta_t$ is \emph{non-decreasing},
$\beta_1\le \beta_2\le \beta_3\le \ldots$;
but it is applied ``in the reverse order'' compared to~\eqref{eq:CBLdiscount}.
So, in both definitions, the older losses are the less weight they
are ascribed.
However, according to~\eqref{eq:CBLdiscount}, the losses $l_t$
have different relative weights in $L_T$, $L_{T+1}$ and so on,
whereas~\eqref{eq:beta-discount} fixes
the relative weight of $l_t$ with respect to all previous losses
forever starting from the moment~$t$.
The latter
property allows us to get uniform algorithms for Learner with
loss guarantees that hold for all $T=1,2,\ldots$;
in contrast, Theorem~2.8 in~\cite{Cesa-BianchiLugosi}
gives a guarantee only at one moment $T$ chosen in advance.
The only kind of discounting that can be expressed
both as~\eqref{eq:CBLdiscount} and as~\eqref{eq:beta-discount}
is the exponential discounting
$\sum_{t=1}^T \alpha^{T-t}l_t$.
Under this discounting, NormalHedge algorithm is
analysed in~\cite{Freund2008};
we briefly compare the obtained bounds in Section~\ref{sec:convex}.

\ifARXIV
Let us say a few words about ``economical'' interpretation of discounting.
Recall that $\alpha_t\le 1$ in Protocol~\ref{prot:GenDisc},
in other words, the previous cumulative loss cannot become
more important at later steps.
If the losses are interpreted as the lost money,
it is more natural to assume that the old losses must be multiplied
by something greater than~$1$.
Indeed, the money could have been invested and have brought some
interest, so the current value of an ancient small loss can 
be considerably large.
Nevertheless, there is 
a not so artificial interpretation for our discounting model as well.
Assume that the loss at each step is expressed 
as a quantity of some goods, and we pay for them in cash;
say, we pay for apples damaged because of 
our incorrect weather prediction.
The price of apples can increase but never decreases.
Then $\beta_t$ in~\eqref{eq:beta-discount} is the current price,
$\sum_{t=1}^T\beta_t l_t$ is the total sum of money we lost,
and $L_T$ is the quantity of apples that we could have bought now
if we had not lost so much money.
(We must also assume that
we cannot hedge our risk by buying 
a lot of cheap apples in advance---the apples 
will rot---and that the bank interest is zero.)

We need the condition $\alpha_t\le 1$
for our algorithms and loss bounds.
However, the case of $\alpha_t\ge 1$ is no less interesting.
We cannot say anything about it and leave it as an open problem,
as well as the general case of arbitrary positive~$\alpha_t$.
\fi

The rest of the paper is organized as follows.
In Section~\ref{sec:linear},
we propose a generalization of
the Aggregating Algorithm~\cite{Vovk:98}
and prove the same bound as in~\cite{Vovk:98}
but for the discounted loss.
In Section~\ref{sec:convex},
we consider convex loss functions
and propose an algorithm similar to
the Weak Aggregating Algotihm~\cite{WAA} and
the exponentially weighted average forecaster
with time-varying learning rate~\cite[\S~2.3]{Cesa-BianchiLugosi},
with a similar loss bound.
In Section~\ref{sec:regression},
we consider the use of prediction with expert advice
for the regression problem
and adapt the
Aggregating Algorithm for Regression~\cite{VovkCOS}
(applied to spaces of linear functions and
to reproducing kernel Hilbert spaces)
to the discounted square loss.
\ifCONF
All our algorithms are inspired by the methodology
of defensive forecasting~\cite{Chernov2010}.
We do not explicitly use or refer to this technique.
However,
to illustrate these ideas we provide
an alternative treatment of the regression task
with the help of defensive forecasting
in the full version
of the paper~\cite{CZ2010} (see Appendix~A.2;
Appendix~A.1 contains some proofs omitted here).
\fi
\ifARXIV
All our algorithms are inspired by the methodology
of defensive forecasting~\cite{Chernov2010}.
We do not explicitly use or refer to this technique
in the main text.
However,
to illustrate these ideas we provide
an alternative treatment of the regression task
with the help of defensive forecasting
in Appendix~\ref{append:DF}.
\fi

\section{Linear Bounds for Learner's Loss}
\label{sec:linear}

In this section, we assume that the set of experts is finite,
$\Theta=\{1,\ldots,K\}$,
and show
how Learner can achieve a bound
of the form $\dL_t\le c\dL_t^k + (c\ln K)/\eta$
for all Experts $k$,
where $c\ge 1$ and $\eta>0$ are constants.
Bounds of this kind were obtained in~\cite{VovkAS}.
Loosely speaking, such a bound holds for certain $c$ and $\eta$
if and only if
the game $(\Omega,\Gamma,\lambda)$
has the following property:
\begin{equation}\label{eq:realiz}
\exists \gamma\in\Gamma \:\forall\omega\in\Omega\quad
\lambda(\gamma,\omega)\le
-\frac{c}{\eta}
  \ln\left(\sum_{i\in I}p_i\e^{-\eta \lambda(\gamma_i,\omega)}\right)
\end{equation}
for any finite index set $I$,
for any $\gamma_i\in\Gamma$, $i\in I$,
and for any $p_i\in[0,1]$ such that $\sum_{i\in I}p_i=1$.
It turns out that this property is sufficient
for the discounted case as well.

\begin{theorem}\label{thm:AAD}
Suppose that the game $(\Omega,\Gamma,\lambda)$
satisfies condition~\eqref{eq:realiz}
for certain $c\ge 1$ and $\eta>0$.
In the game played according to Protocol~\ref{prot:GenDisc},
Learner has a strategy guaranteeing
that, for any $T$ and for any $k\in\{1,\ldots, K\}$, it holds
\begin{equation}\label{eq:AAbound}
   \dL_T \le c \dL_T^k + \frac{c\ln K}{\eta}\,.
\end{equation}
\end{theorem}
\ifARXIV
We formulate the strategy for Learner in Subsection~\ref{ssec:AAD}
and prove the theorem in Subsection~\ref{ssec:AAproof}.
\fi

For the standard undiscounted case
(Accountant announces $\alpha_t=1$ at each step $t$),
this theorem was proved by Vovk
in~\cite{VovkAS} with the help of the Aggregating Algorithm (AA)
as Learner's strategy.
It is known (\cite{Haussler1998,Vovk:98})
that this bound is asymptotically optimal
for large pools of Experts
(for games satisfying some assumptions):
if the game does not satisfy~\eqref{eq:realiz}
for some $c\ge 1$ and $\eta>0$,
then, for sufficiently large $K$,
there is a strategy for Experts and Reality
(recall that Accountant always says $\alpha_t=1$)
such that Learner cannot secure~\eqref{eq:AAbound}.
For the special case of $c=1$,
bound~\eqref{eq:AAbound} is tight
for any fixed $K$ as well~\cite{Vovk:1999derandomizing}.
These results imply optimality of Theorem~\ref{thm:AAD}
in the new setting with general discounting
(when we allow arbitrary behaviour of Accountant
with the only requirement $\alpha_t\in(0,1]$).
However,
they leave open the question of lower bounds
under different discounting assumptions
(that is, when Accountant moves are fixed);
a particularly interesting case is
the exponential discounting
$\alpha_t=\alpha\in(0,1)$.

\ifCONF
\vspace*{-10ex}
\begin{proof}
\fi
\ifARXIV
\subsection{Learner's Strategy}
\label{ssec:AAD}
\fi

\ifARXIV
To prove Theorem~\ref{thm:AAD},
we will exploit the AA with a minor modification.

\begin{algorithm}
  \caption{The Aggregating Algorithm}\label{alg:AA}
  \begin{algorithmic}[1]
   \STATE Initialize weights of Experts $w_0^k:=1/K$, $k=1,\ldots,K$.
   \FOR{$t=1,2,\dots$}
     \STATE Get Experts' predictions
                             {$\gamma_t^k \in \Gamma, k=1,\ldots,K$}.
     \STATE Calculate $g_t(\omega)=
            -\frac{c}{\eta}\ln\left(\sum_{k=1}^K w_{t-1}^k
                \e^{-\eta \lambda(\gamma_t^k,\omega)}\right)$,
                   for all $\omega \in \Omega$.
     \STATE Output $\gamma_t := \sigma(g_t) \in \Gamma$.
     \STATE Get $\omega_t\in\Omega$.
     \STATE\label{AAline:update} Update the weights
            $\tilde w_t^k := w_{t-1}^k \e^{-\eta \lambda(\gamma_t^k,\omega_t)}$,
               $k=1,\ldots,K$,
     \STATE\label{AAline:normal} and normalize them
            $w_t^k := \tilde w_t^k / \sum_{k=1}^K \tilde w_t^k$, $k=1,\ldots,K$.
   \ENDFOR.
  \end{algorithmic}
\end{algorithm}
The pseudocode of the AA is given as Algorithm~\ref{alg:AA}.
The algorithm has three parameters,
which depend on the game $(\Omega,\Gamma,\lambda)$:
$c\ge 1$, $\eta>0$, and a function $\sigma\colon\R^\Omega\to\Gamma$.
The function $\sigma$ is called a \emph{substitution function}
and must have the following property:
$\lambda(\sigma(g),\omega)\le g(\omega)$ for all $\omega\in\Omega$
if for $g\in\R^\Omega$ there exists any $\gamma\in\Gamma$ such that
$\lambda(\gamma,\omega)\le g(\omega)$ for all $\omega\in\Omega$.
A natural example of substitution function is given by
\begin{equation}\label{eq:subst}
\sigma(g)=
   \arg\min_{\gamma\in\Gamma}\bigl(\lambda(\gamma,\omega) - g(\omega)\bigr)
\end{equation}
(if the minimum is attained at several points,
one can take any of them).
An advantage of this $\sigma$ is that the normalization step
in line~\ref{AAline:normal}
is not necessary and one can take $w_t^k=\tilde w_t^k$.
Indeed, multiplying all $w_t^k$ by a constant (independent of $k$)
we add to all $g_t(\omega)$ a constant (independent of $\omega$),
and $\sigma(g_t)$ does not change.

The Aggregating Algorithm with Discounting (AAD)
differs only by the use of the weights
in the computation of $g_t$
and the update of the weights.

The pseudocode of the AAD is given as Algorithm~\ref{alg:AAD}.
\fi
\ifCONF
As Learner's strategy we exploit a minor modification of
the Aggregating Algorithm, the AA with Discounting (AAD).
The pseudocode is given as Algorithm~\ref{alg:AAD}.
\fi

\begin{algorithm}
  \caption{The Aggregating Algorithm with Discounting}\label{alg:AAD}
  \begin{algorithmic}[1]
   \STATE Initialize weights of Experts $w_0^k:=1$, $k=1,\ldots,K$.
   \FOR{$t=1,2,\dots$}
     \STATE Get discount $\alpha_{t-1}\in(0,1]$.
     \STATE Get Experts' predictions
                             {$\gamma_t^k \in \Gamma, k=1,\ldots,K$}.
     \STATE Calculate $g_t(\omega)=
            -\frac{c}{\eta}\left(\ln\sum_{k=1}^K
                \frac{1}{K}(w_{t-1}^k)^{\alpha_{t-1}}
                \e^{-\eta \lambda(\gamma_t^k,\omega)}\right)$,
                   for all $\omega \in \Omega$.
     \STATE Output $\gamma_t := \sigma(g_t) \in \Gamma$.
     \STATE Get $\omega_t\in\Omega$.
     \STATE\label{AADline:update} Update the weights
            $w_t^k := (w_{t-1}^k)^{\alpha_{t-1}}
                    \e^{\eta \lambda(\gamma_t,\omega_t)/c
                       -\eta \lambda(\gamma_t^k,\omega_t)}$,
               $k=1,\ldots,K$,
   \ENDFOR.
  \end{algorithmic}
\end{algorithm}
\ifCONF
The algorithm has three parameters,
which depend on the game $(\Omega,\Gamma,\lambda)$:
$c\ge 1$, $\eta>0$, and a function $\sigma\colon\R^\Omega\to\Gamma$.
The function $\sigma$ is called a \emph{substitution function}
and must have the following property:
$\lambda(\sigma(g),\omega)\le g(\omega)$ for all $\omega\in\Omega$
if for $g\in\R^\Omega$ there exists any $\gamma\in\Gamma$ such that
$\lambda(\gamma,\omega)\le g(\omega)$ for all $\omega\in\Omega$.
A natural example of substitution function is given by
\begin{equation}\label{eq:subst}
\sigma(g)=
   \arg\min_{\gamma\in\Gamma}\bigl(\lambda(\gamma,\omega) - g(\omega)\bigr)
\end{equation}
(if the minimum is attained in several points,
one can take any of them).
An advantage of this $\sigma$
is that one can use in line~\ref{AADline:update} of the algorithm
the update rule
$w_t^k := (w_{t-1}^k)^{\alpha_{t-1}}
     \e^{-\eta \lambda(\gamma_t^k,\omega_t)}$,
which does not contain Learner's losses.
Indeed, multiplying all $w_t^k$ by a constant (independent of $k$)
we add to all $g_t(\omega)$ a constant (independent of $\omega$),
and $\sigma(g_t)$ does not change.
\fi
\ifARXIV
For a substitution function satisfying~\eqref{eq:subst},
one can use in line~\ref{AADline:update} the update rule
$w_t^k := (w_{t-1}^k)^{\alpha_{t-1}}
     \e^{-\eta \lambda(\gamma_t^k,\omega_t)}$,
which does not contain Learner's losses,
in the same manner as
the normalization in Algorithm~\ref{alg:AA}
can be omitted.
%
%
\fi

\ifARXIV
\subsection{Proof of the Bound}
\label{ssec:AAproof}
\fi

Assume that $c$ and $\eta$ are such that
condition~\eqref{eq:realiz} holds for the game.
Let us show that
Algorithm~\ref{alg:AAD}
preserves the following condition:
\begin{equation}\label{eq:AADcond}
\sum_{k=1}^K \frac{1}{K}w_t^k\le 1\,.
\end{equation}
Condition~\eqref{eq:AADcond} trivially holds for $t=0$.
Assume that \eqref{eq:AADcond}~holds for $t-1$,
that is, $\sum_{k=1}^K w_{t-1}^k/K\le 1$.
Thus, we have
\ifCONF
$
\sum_{k=1}^K (w_{t-1}^k)^{\alpha_{t-1}}/K
\le \left(\sum_{k=1}^K w_{t-1}^k/K\right)^{\alpha_{t-1}}
\le 1
$,
\fi
\ifARXIV
$$
\sum_{k=1}^K \frac{1}{K}(w_{t-1}^k)^{\alpha_{t-1}}
\le \left(\sum_{k=1}^K \frac{1}{K}w_{t-1}^k\right)^{\alpha_{t-1}}
\le 1\,,
$$
\fi
since the function $x\mapsto x^\alpha$ is concave
for $\alpha\in(0,1]$, $x\ge 0$,
and since $x\le 1$ implies $x^\alpha\le 1$ for $\alpha\ge 0$
and $x\ge 0$.

Let $\tilde w^k$ be any reals such that
$\tilde w^k\ge (w_{t-1}^k)^{\alpha_{t-1}}/K$
and $\sum_{k=1}^K\tilde w^k = 1$.
Due to condition~\eqref{eq:realiz}
there exists $\gamma\in\Gamma$
such that for all $\omega\in\Omega$
\begin{multline*}
\lambda(\gamma,\omega)
\le
-\frac{c}{\eta}
  \ln\left(\sum_{k=1}^K \tilde w^k\e^{-\eta \lambda(\gamma_t^k,\omega)}\right)
\\
\le
-\frac{c}{\eta}
  \ln\left(\sum_{k=1}^K \frac{1}{K}(w_{t-1}^k)^{\alpha_{t-1}}
                  \e^{-\eta \lambda(\gamma_t^k,\omega)}\right)
= g_t(\omega)
\end{multline*}
(the second inequality holds due to our choice of $\tilde w^k$).
Thus, due to the property of $\sigma$,
we have $\lambda(\gamma_t,\omega)\le g_t(\omega)$ for all $\omega\in\Omega$.
In particular, this holds for $\omega=\omega_t$,
and we get%
\ifCONF~\eqref{eq:AADcond}.
\fi
\ifARXIV
$$
\lambda(\gamma_t,\omega_t)
\le
-\frac{c}{\eta}
  \ln\left(\sum_{k=1}^K \frac{1}{K}(w_{t-1}^k)^{\alpha_{t-1}}
                  \e^{-\eta \lambda(\gamma_t^k,\omega_t)}\right)\,,
$$
which is equivalent to~\eqref{eq:AADcond}.

\fi
To get the loss bound~\eqref{eq:AAbound}, it remains to note that
$$
\ln w_t^k = \eta\left(\dL_t/c - \dL_t^k\right)\,.
$$
Indeed, for $t=0$, this is trivial.
If this holds for $w_{t-1}^k$,
then
\ifCONF
$\ln w_t^k =
{\alpha_{t-1}}\ln (w_{t-1}^k)
+\eta \lambda(\gamma_t,\omega_t)/c -\eta \lambda(\gamma_t^k,\omega_t)
=
\alpha_{t-1}\eta\left(\dL_{t-1}/c - \dL_{t-1}^k\right)
+\eta \lambda(\gamma_t,\omega_t)/c -\eta \lambda(\gamma_t^k,\omega_t)
=
\eta\left((\alpha_{t-1}\dL_{t-1} + \lambda(\gamma_t,\omega_t))/c -
          (\alpha_{t-1}\dL_{t-1}^k + \lambda(\gamma_t^k,\omega_t))
    \right)
=\eta\left(\dL_t/c - \dL_t^k\right)
$
\fi
\ifARXIV
\begin{multline*}
\ln w_t^k =
{\alpha_{t-1}}\ln (w_{t-1}^k)
+\eta \lambda(\gamma_t,\omega_t)/c -\eta \lambda(\gamma_t^k,\omega_t)
\\
=
\alpha_{t-1}\eta\left(\dL_{t-1}/c - \dL_{t-1}^k\right)
+\eta \lambda(\gamma_t,\omega_t)/c -\eta \lambda(\gamma_t^k,\omega_t)
\\
=
\eta\left((\alpha_{t-1}\dL_{t-1} + \lambda(\gamma_t,\omega_t))/c -
          (\alpha_{t-1}\dL_{t-1}^k + \lambda(\gamma_t^k,\omega_t))
    \right)
=\eta\left(\dL_t/c - \dL_t^k\right)
\end{multline*}
\fi
and we get the equality for $w_t^k$.
Thus, condition~\eqref{eq:AADcond} means that
\begin{equation}\label{eq:AADFbound}
\sum_{k=1}^K \frac{1}{K} \e^{\eta\left(\dL_t/c - \dL_t^k\right)}\le 1\,,
\end{equation}
and~\eqref{eq:AAbound} follows
by lower-bounding the sum by any of its terms.
\ifCONF
\EP
\fi

\begin{remark}
Everything in this section remains valid,
if we replace the equal initial Experts' weights $1/K$
by arbitrary non-negative weights $w^k$, $\sum_{k=1}^K w^k=1$.
This leads to a variant of~\eqref{eq:AAbound},
where the last additive term is replaced by
$\frac{c}{\eta}\ln\frac{1}{w^k}$.
Additionally, we can consider any measurable space $\Theta$
of Experts and a non-negative weight function $w(\theta)$,
and replace sums over $K$ by integrals over $\Theta$.
Then the algorithm and its analysis remain valid
(if we impose natural integrability conditions
on Experts' predictions $\gamma_t^\theta$;
see~\cite{VovkCOS} for more detailed discussion)---this will
be used in Section~\ref{sec:regression}.
\end{remark}


\section{Learner's Loss in Bounded Convex Games}
\label{sec:convex}

The linear bounds of the form~\eqref{eq:AAbound}
are perfect when $c=1$.
However, for many games (for example, the absolute loss game),
condition~\eqref{eq:realiz}
does not hold for $c=1$ (with any $\eta>0$),
and one cannot get a bound of the form $\dL_t\le \dL_t^k + O(1)$.
Since Experts' losses $\dL_T^\theta$ may grow
as $T$ in the worst case,
any bound with $c>1$ only guarantees that
Learner's loss may exceed an Expert's loss
by at most $O(T)$.
However, for a large class of interesting games
(including the absolute loss game),
one can obtain guarantees of the form
$\dL_T\le \dL_T^k + O(\sqrt{T})$
in the undiscounted case.
In this section, we prove an analogous result
for the discounted setting.

A game $(\Omega,\Gamma,\lambda)$ is non-empty
if $\Omega$ and $\Gamma$ are non-empty.
The game is called \emph{bounded}
if $L=\max_{\omega,\gamma} \lambda(\gamma,\omega)<\infty$.
One may assume that $L=1$
(if not, consider the scaled loss function $\lambda/L$).
The game is called \emph{convex} if
for any predictions $\gamma_1,\ldots,\gamma_M\in\Gamma$ and
for any weights $p_1,\ldots,p_M\in[0,1]$, ${\sum_{m=1}^M p_m=1}$,
\ifCONF
there exists $\gamma\in\Gamma$ such that
$\lambda(\gamma,\omega)\le\sum_{m=1}^M p_m\lambda(\gamma_m,\omega)$
for all $\omega\in\Omega$.
\fi
\ifARXIV
\begin{equation}\label{eq:convexity}
\exists\gamma\in\Gamma\:\forall\omega\in\Omega\quad
\lambda(\gamma,\omega)\le\sum_{m=1}^M p_m\lambda(\gamma_m,\omega)\,.
\end{equation}
\fi
Note that if $\Gamma$ is a convex set (e.\,g.,\,$\Gamma=[0,1]$)
and $\lambda(\gamma,\omega)$ is convex in $\gamma$
(e.\,g.,\,$\lambda^\mathrm{abs}$),
then the game is convex.

\begin{theorem}\label{thm:convexbound}
Suppose that $(\Omega,\Gamma,\lambda)$
is a non-empty convex game,
and $\lambda(\gamma,\omega)\in[0,1]$
for all $\gamma\in\Gamma$ and $\omega\in\Omega$.
In the game played according to Protocol~\ref{prot:GenDisc},
Learner has a strategy guaranteeing
that, for any $T$ and for any $k\in\{1,\ldots, K\}$, it holds
\begin{equation}\label{eq:sqrtbound}
\dL_T \le \dL_T^k + \sqrt{\ln K}\sqrt{\frac{B_T}{\beta_T}}\,,
\end{equation}
where $\beta_t=1/(\alpha_1\cdots\alpha_{t-1})$ and
$B_T=\sum_{t=1}^T \beta_t$.
\end{theorem}

Note that $B_T/\beta_T$ is the maximal predictors' loss,
which incurs when the predictor suffers the maximal
possible loss $l_t=1$ at each step.
\ifARXIV

\fi
In the undiscounted case, $\alpha_t=1$,
thus $\beta_t=1$, $B_T=T$,
and~\eqref{eq:sqrtbound} becomes
\ifCONF$\fi
\ifARXIV$$\fi
\dL_T \le \dL_T^k + \sqrt{T\ln K}%
\ifCONF$. \fi
\ifARXIV\,.$$\fi
A similar bound
(but with worse constant $\sqrt{2}$ instead of $1$
before $\sqrt{T\ln K}$) is
obtained in~\cite[Theorem~2.3]{Cesa-BianchiLugosi}:
\ifCONF$\dL_T \le \dL_T^k + \sqrt{2T\ln K} + \sqrt{(\ln K)/8}$. \fi
\ifARXIV
$$
\dL_T \le \dL_T^k + \sqrt{2T\ln K} + \sqrt{\frac{\ln K}{8}}\,.
$$

\fi
For the exponential discounting $\alpha_t=\alpha$,
we have $\beta_t=\alpha^{-t+1}$ and $B_T=(1-\alpha^{-T})/(1-1/\alpha)$,
and \eqref{eq:sqrtbound} transforms into
\ifCONF
$
\dL_T \le \dL_T^k + \sqrt{\ln K}\sqrt{(1-\alpha^T)/(1-\alpha)}
\le
\dL_T^k + \sqrt{(\ln K)/(1-\alpha)}
$.
\fi
\ifARXIV
$$
\dL_T \le \dL_T^k + \sqrt{\ln K}\sqrt{\frac{1-\alpha^T}{1-\alpha}}
\le
\dL_T^k + \sqrt{\frac{\ln K}{1-\alpha}}\,.
$$
\fi
A similar bound (with worse constants) is obtained
in~\cite{Freund2008} for NormalHedge:
\ifCONF
$
\dL_T
\le
\dL_T^k + \sqrt{(8\ln 2.32K)/(1-\alpha)}
$.
\fi
\ifARXIV
$$
\dL_T
\le
\dL_T^k + \sqrt{\frac{8\ln 2.32K}{1-\alpha}}\,.
$$
\fi
The NormalHedge algorithm has an important advantage:
it can guarantee the last bound without knowledge
of the number of experts~$K$
(see~\cite{CFH2009} for a precise definition).
We can achieve the same with the help of a more complicated
algorithm but at the price of a worse bound 
\ifCONF (see also the remark after the proof).\fi
\ifARXIV (Theorem~\ref{thm:convexsuperbound}).\fi

\ifCONF
\begin{proof}
\fi
\ifARXIV
\subsection{Learner's Strategy for Theorem~\ref{thm:convexbound}}
\fi
The pseudocode of Learner's strategy
is given as Algorithm~\ref{alg:FDFD}.
It contains a constant $a>0$, which we will choose later in the proof.

\ifARXIV
The algorithm is not fully specified,
since lines~\ref{FDFDline:find}--\ref{FDFDline:gamma}
of Algorithm~\ref{alg:FDFD} 
allow arbitrary choice of $\gamma$ satisfying the inequality.
The algorithm can be completed with the help of 
a substitution function $\sigma$
as in Algorithm~\ref{alg:AAD},
so that lines~\ref{FDFDline:find}--\ref{FDFDline:output}
are replaced by
$$g_t(\omega)=
 -\frac{1}{\eta_t}\ln\left(
          \sum_{k=1}^K \frac{1}{K}
                 \left(w_{t-1}^k\right)^{\alpha_{t-1}\eta_t/\eta_{t-1}}
                             \e^{
                                 -\eta_t\lambda(\gamma_t^k,\omega)
                                   -\eta_t^2/8
                                 }
                 \right)
$$
and $\gamma_t=\sigma(g_t)$.
However, the current form of Algorithm~\ref{alg:FDFD}
emphasizes the similarity to the Algorithm~\ref{alg:fullFDFD},
which is described later (Subsection~\ref{ssec:epsilonbest})
but actually inspired our analysis.

\fi

\begin{algorithm}[ht]
  \caption{Learner's Strategy for Convex Games}
\label{alg:FDFD}
  \begin{algorithmic}[1]
   \STATE Initialize weights of Experts $w_0^k:=1$, $k=1,\ldots,K$.\\
          Set $\beta_1=1$, $B_0=0$.
   \FOR{$t=1,2,\dots$}
     \STATE Get discount $\alpha_{t-1}\in(0,1]$; 
            update $\beta_t=\beta_{t-1}/\alpha_{t-1}$, $B_t=B_{t-1}+\beta_t$.
     \STATE Compute $\eta_t=a\sqrt{\beta_t/B_t}$.
     \STATE Get Experts' predictions
                             {$\gamma_t^k \in \Gamma$, $k=1,\ldots,K$}.
     \STATE \label{FDFDline:find}
            Find $\gamma\in\Gamma$ s.t. for all $\omega\in\Omega$
     \STATE \label{FDFDline:gamma} \qquad
            $\lambda(\gamma,\omega)
              \le
             -\frac{1}{\eta_t}\ln\left(
                        \sum_{k=1}^K \frac{1}{K}
                             \left(w_{t-1}^k\right)^{\alpha_{t-1}\eta_t/\eta_{t-1}}
                             \e^{
                                 -\eta_t\lambda(\gamma_t^k,\omega)
                                   -\eta_t^2/8
                                 }
                                 \right)
            $
     \STATE \label{FDFDline:output} Output $\gamma_t := \gamma$.
     \STATE Get $\omega_t\in\Omega$.
     \STATE \label{FDFDline:update}
            Update the weights
            $w_t^k := \left(w_{t-1}^k\right)^{\alpha_{t-1}\eta_t/\eta_{t-1}}
                             \e^{
                                 \eta_t\bigl(\lambda(\gamma_t,\omega_t)
                                        -\lambda(\gamma_t^k,\omega_t)\bigr)
                                   -\eta_t^2/8
                                 }
            $,
     \STATE \qquad          $k=1,\ldots,K$,
   \ENDFOR.
  \end{algorithmic}
\ifCONF
  \textbf{Remark:}

  If $\lambda(\gamma,\omega)$ is convex in $\gamma$,
  lines~\ref{FDFDline:find}--\ref{FDFDline:gamma}
  can be replaced by
  $\gamma=\sum_{k=1}^K\tilde w^k \gamma_t^k$, see~\eqref{eq:gammaexists}.
\fi
\end{algorithm}

\ifARXIV

Let us explain the relation of Algorithm~\ref{alg:FDFD}
to the Weak Aggregating Algorithm~\cite{WAA}
and the exponentially weighted average forecaster
with time-varying learning rate~\cite[\S~2.3]{Cesa-BianchiLugosi}.
To this end, consider Algorithm~\ref{alg:WAAD}.

\begin{algorithm}[ht]
  \caption{Weak Aggregating Algorithm with Discounting}
\label{alg:WAAD}
  \begin{algorithmic}[1]
   \STATE Initialize Experts' cumulative losses $\dL_0^k:=0$, $k=1,\ldots,K$.\\
          Set $\beta_1=1$, $B_0=0$.
   \FOR{$t=1,2,\dots$}
     \STATE Get discount $\alpha_{t-1}\in(0,1]$; 
            update $\beta_t=\beta_{t-1}/\alpha_{t-1}$, $B_t=B_{t-1}+\beta_t$.
     \STATE Compute $\eta_t=a\sqrt{\beta_t/B_t}$.
     \STATE Compute the weights 
               $q_t^k=\e^{-\alpha_{t-1}\eta_t \dL_{t-1}^k}$, $k=1,\ldots,K$.
     \STATE Compute the normalized weights
               $\tilde w_t^k = q_t^k\left/\sum_{j=1}^K q_t^j\right.$.
     \STATE Get Experts' predictions
                             {$\gamma_t^k \in \Gamma$, $k=1,\ldots,K$}.
     \STATE \label{WAADline:find}
            Find $\gamma\in\Gamma$ s.t. for all $\omega\in\Omega$
            \quad 
            $\lambda(\gamma,\omega)\le 
             \sum_{k=1}^K \tilde w_t^k\lambda(\gamma_t^k,\omega)
            $.
     \STATE \label{WAADline:output} Output $\gamma_t := \gamma$.
     \STATE Get $\omega_t\in\Omega$.
     \STATE Update $\dL_t^k:=\alpha_{t-1}\dL_{t-1}^k+\lambda(\gamma_t^k,\omega_t)$,
                            $k=1,\ldots,K$.
   \ENDFOR.
  \end{algorithmic}
\end{algorithm}
The proof of Theorem~\ref{thm:convexbound} 
implies that Algorithm~\ref{alg:WAAD} is 
a special case of Algorithm~\ref{alg:FDFD}.
Indeed, \eqref{eq:FDFDweights} implies that 
$w_{t-1}^k = \e^{-\eta_{t-1}\dL_{t-1}^k + C}$,
where $C$ does not depend on~$k$
and $w_{t-1}^k$ are the weights from Algorithm~\ref{alg:FDFD}.
Therefore $q_t^k = C'(w_{t-1}^k)^{\alpha_{t-1}\eta_t/\eta_{t-1}}$,
where $C'$ does not depend on~$k$,
and one can take $\tilde w_t^k$ for $\tilde w^k$ in the proof
of Theorem~\ref{thm:convexbound}.
Thus, if Algorithm~\ref{alg:WAAD} output some $\gamma_t$
then Algorithm~\ref{alg:FDFD} can output this $\gamma_t$ as well.

Recall that if $\alpha_t=1$ for all $t$ (the undiscounted case),
$\beta_t=1$ and $B_t=t$, hence $\eta_t=a/\sqrt{t}$.
In this case, Algorithm~\ref{alg:WAAD} is just 
the Weak Aggregating Algorithm as described in~\cite{WAA}.

Consider now the case when 
$\Gamma$ is a convex set and 
$\lambda(\gamma,\omega)$ is convex in $\gamma$.
Then one can take 
$\gamma_t = \sum_{k=1}^K \tilde w_t^k \gamma_t^k$
in Algorithm~\ref{alg:WAAD}.
For $\alpha_t=1$, we get exactly
the exponentially weighted average forecaster
with time-varying learning rate~\cite[\S~2.3]{Cesa-BianchiLugosi}.

\subsection{Proof of Theorem~\ref{thm:convexbound}}
\fi

Similarly to the case of the AAD,
let us show that Algorithm~\ref{alg:FDFD}
always can find $\gamma$ in lines~\ref{FDFDline:find}--\ref{FDFDline:gamma}
and preserves the following condition:
\begin{equation}\label{eq:FDFDcond}
\sum_{k=1}^K \frac{1}{K}w_t^k\le 1\,.
\end{equation}

First check that $\alpha_{t-1}\eta_t/\eta_{t-1}\le 1$.
Indeed, $\alpha_{t-1}=\beta_{t-1}/\beta_t$,
and thus
\begin{equation}\label{eq:alphaless1}
\alpha_{t-1}\frac{\eta_t}{\eta_{t-1}}
=
 \frac{\beta_{t-1}}{\beta_t}
 \frac{a\sqrt{\beta_t/B_t}}{a\sqrt{\beta_{t-1}/B_{t-1}}}
=\sqrt{\frac{\beta_{t-1}}{\beta_t}\frac{B_{t-1}}{B_t}}
=\sqrt{\alpha_{t-1}}\sqrt{\frac{B_{t-1}}{B_{t-1}+\beta_{t}}}
\le 1\,.
\end{equation}

Condition~\eqref{eq:FDFDcond} trivially holds for $t=0$.
Assume that \eqref{eq:FDFDcond}~holds for $t-1$,
that is, $\sum_{k=1}^K w_{t-1}^k/K\le 1$.
Thus, we have
\begin{equation}\label{eq:FDFDweightsSeminormal}
\sum_{k=1}^K \frac{1}{K}(w_{t-1}^k)^{\alpha_{t-1}\eta_t/\eta_{t-1}}
\le \left(\sum_{k=1}^K \frac{1}{K}w_{t-1}^k\right)^{\alpha_{t-1}\eta_t/\eta_{t-1}}
\le 1\,,
\end{equation}
since the function $x\mapsto x^\alpha$ is concave
for $\alpha\in(0,1]$, $x\ge 0$,
and since $x\le 1$ implies $x^\alpha\le 1$ for $\alpha\ge 0$
and $x\ge 0$.

Let $\tilde w^k$ be any reals such that
$\tilde w^k\ge (w_{t-1}^k)^{\alpha_{t-1}\eta_t/\eta_{t-1}}/K$
and ${\sum_{k=1}^K\tilde w^k = 1}$.
(For example, 
$\tilde w^k = (w_{t-1}^k)^{\alpha_{t-1}\eta_t/\eta_{t-1}}
 \left/
  \sum_{j=1}^K (w_{t-1}^j)^{\alpha_{t-1}\eta_t/\eta_{t-1}}
 \right.
$.)
By the Hoeffding inequality (see, e.\,g., \cite[Lemma~2.2]{Cesa-BianchiLugosi}),
we have
\begin{equation}\label{eq:Hoeff}
\ln \sum_{k=1}^K\tilde w^k  \e^{-\eta_t\lambda(\gamma_t^k,\omega)}
\le
-\eta_t \sum_{k=1}^K\tilde w^k \lambda(\gamma_t^k,\omega) + \frac{\eta_t^2}{8}\,,
\end{equation}
since $\lambda(\gamma,\omega)\in[0,1]$
for any $\gamma\in\Gamma$ and $\omega\in\Omega$.
Since the game is convex,
there exists $\gamma\in\Gamma$ such that
$\lambda(\gamma,\omega)\le \sum_{k=1}^K\tilde w^k \lambda(\gamma_t^k,\omega)$
for all $\omega\in\Omega$.
For this $\gamma$ and for all $\omega\in\Omega$ we have
\begin{multline}\label{eq:gammaexists}
\lambda(\gamma,\omega)
\le
\sum_{k=1}^K\tilde w^k \lambda(\gamma_t^k,\omega)
\le
-\frac{1}{\eta_t}
  \ln\left(\sum_{k=1}^K \tilde w^k
            \e^{-\eta \lambda(\gamma_t^k,\omega) - \eta_t^2/8}
     \right)
\\
\le
-\frac{1}{\eta_t}
  \ln\left(\sum \frac{1}{K}
           \left(w_{t-1}^k\right)^{\alpha_{t-1}\eta_t/\eta_{t-1}}
           \e^{-\eta_t\lambda(\gamma_t^k,\omega)-\eta_t^2/8}
     \right)
\end{multline}
(the second inequality follows from~\eqref{eq:Hoeff}, and
the third inequality holds due to our choice of $\tilde w^k$).
Thus, one can always find
$\gamma$ in lines~\ref{FDFDline:find}--\ref{FDFDline:gamma}
of Algorithm~\ref{alg:FDFD}.
It remains to note that the inequality in line~\ref{FDFDline:gamma}
with $\gamma_t$ substituted for $\gamma$
and $\omega_t$ substituted for $\omega$
is equivalent to
$$
1\ge \sum \frac{1}{K}
           \left(w_{t-1}^k\right)^{\alpha_{t-1}\eta_t/\eta_{t-1}}
           \e^{\eta_t\lambda(\gamma_t,\omega_t)
              -\eta_t\lambda(\gamma_t^k,\omega_t)-\eta_t^2/8}
=
     \sum \frac{1}{K} w_{t}^k\,.
$$

Now let us check that
\begin{equation}\label{eq:FDFDweights}
\ln w_t^k = \eta_t\left(\dL_t - \dL_t^k\right)
             -\frac{\eta_t}{8\beta_t}\sum_{\tau=1}^t \beta_\tau\eta_\tau
\,.
\end{equation}
Indeed, for $t=0$, this is trivial.
Assume that it holds for $w_{t-1}^k$. Then,
taking the logarithm of the update expression
in line~\ref{FDFDline:update} \ifARXIV of Algorithm~\ref{alg:FDFD} \fi
and substituting $\ln w_{t-1}^k$,
we get
\begin{multline*}\allowdisplaybreaks
\ln w_t^k
=
\frac{\alpha_{t-1}\eta_t}{\eta_{t-1}}\ln w_{t-1}^k
+ \eta_t\bigl(\lambda(\gamma_t,\omega_t)
 -\lambda(\gamma_t^k,\omega_t)\bigr)
 -\frac{\eta_t^2}{8}
\ifCONF=\fi
\\
=
\frac{\alpha_{t-1}\eta_t}{\eta_{t-1}}
\left(
     \eta_{t-1}\left(\dL_{t-1} - \dL_{t-1}^k\right)
    -\frac{\eta_{t-1}}{8\beta_{t-1}}\sum_{\tau=1}^{t-1} \beta_\tau\eta_\tau
\right)
+ \eta_t\bigl(\lambda(\gamma_t,\omega_t)
 -\lambda(\gamma_t^k,\omega_t)\bigr)
 -\frac{\eta_t^2}{8}
\\
=
\eta_t\left(\alpha_{t-1}\dL_{t-1}+\lambda(\gamma_t,\omega_t)
           -\alpha_{t-1}\dL_{t-1}^k-\lambda(\gamma_t^k,\omega_t)
      \right)
-\frac{\eta_t}{8\beta_t}\sum_{\tau=1}^{t-1} \beta_\tau\eta_\tau
-\frac{\eta_t^2}{8}
\\
=
\eta_t\left(\dL_t - \dL_t^k\right)
-\frac{\eta_t}{8\beta_t}\sum_{\tau=1}^t \beta_\tau\eta_\tau\,.
\end{multline*}

Condition~\eqref{eq:FDFDcond} implies that
$w_T^k\le K$ for all $k$ and $T$,
hence we get a loss bound
\begin{equation}\label{eq:preFDFDloss}
\dL_T \le \dL_T^k + \frac{\ln K}{\eta_T} +
   \frac{1}{8\beta_T}\sum_{t=1}^T \beta_t\eta_t\,.
\end{equation}

Recall that $\eta_t=a\sqrt{\beta_t/B_t}$.
To estimate $\sum_{t=1}^T \beta_t\eta_t$,
we use the following inequality
\ifCONF
(see~\cite[Appendix~A.1]{CZ2010} for the proof).
\fi
\ifARXIV
(see Appendix~\ref{append:technical} for the proof).
\fi

\begin{lemma}\label{lem:gen-sqrt-sum}
Let $\beta_t$ be any reals such that
$1\le\beta_1\le\beta_2\le\ldots$.
Let $B_T=\sum_{t=1}^T\beta_t$.
Then, for any $T$, it holds
$$
  \frac{1}{\beta_T}\sum_{t=1}^T \beta_t\sqrt{\frac{\beta_t}{B_t}}
  \le
  2\sqrt{\frac{B_T}{\beta_T}}\,.
$$
\end{lemma}
Then~\eqref{eq:preFDFDloss} implies
$$
\dL_T \le \dL_T^k + \frac{\ln K}{a}\sqrt{\frac{B_T}{\beta_T}} +
   \frac{2a}{8}\sqrt{\frac{B_T}{\beta_T}}
=
\dL_T^k + \left(\frac{\ln K}{a} + \frac{a}{4}
          \right)\sqrt{\frac{B_T}{\beta_T}}
\,.
$$
Choosing $a=2\sqrt{\ln K}$,
we finally get
\ifCONF
the bound.
\fi
\ifARXIV
$$
\dL_T \le \dL_T^k + \sqrt{\ln K}\sqrt{\frac{B_T}{\beta_T}}\,.
$$
\fi
\ifCONF
\EP

\begin{remark}
Algorithm~\ref{alg:FDFD} is a modification
of the ``Fake Defensive Forecasting'' algorithm
from~\cite[Theorem~9]{CV2010}.
The algorithm is based on the ideas of defensive 
forecasting~\cite{Chernov2010},
in particular, Hoeffding supermartingales~\cite{Vovk-Hoeffding},
combined with the ideas from an early version
of the Weak Aggregating Algorithm~\cite{WAAreport}.
However, our analysis is completely different from~\cite{CV2010},
following the lines of~\cite[Theorem~2.2]{Cesa-BianchiLugosi}
and~\cite{WAAreport}.
Algorithm~\ref{alg:FDFD} is quite similar to
the exponentially weighted average forecaster
with time-varying learning rate~\cite[\S~2.3]{Cesa-BianchiLugosi},
but it keeps the weights $w_k^t/K$ semi-normalized
because of a specific update rule in line~\ref{FDFDline:update}
instead of normalizing them.
A more involved version of Algorithm~\ref{alg:FDFD}
can achieve a bound for $\epsilon$-quantile regret~\cite{CFH2009},
but the analysis becomes more complicated,
requires application of the supermartingale technique,
and gives a worse bound.
\end{remark}
\fi

\ifARXIV

\subsection{A Bound with respect to $\epsilon$-Best Expert}
\label{ssec:epsilonbest}

Algorithm~\ref{alg:FDFD} originates in
the ``Fake Defensive Forecasting'' (FDF) algorithm
from~\cite[Theorem~9]{CV2010}.
That algorithm is based on the ideas of defensive 
forecasting~\cite{Chernov2010},
in particular, Hoeffding supermartingales~\cite{Vovk-Hoeffding},
combined with the ideas from an early version
of the Weak Aggregating Algorithm~\cite{WAAreport}.
However, our analysis in Theorem~\ref{thm:convexbound} 
is completely different from~\cite{CV2010},
following the lines of~\cite[Theorem~2.2]{Cesa-BianchiLugosi}
and~\cite{WAAreport}.

In this subsection,
we consider a direct extension of
the FDF algorithm from~\cite[Theorem~9]{CV2010}
to the discounted case.
Algorithm~\ref{alg:fullFDFD}
becomes the FDF algorithm when $\alpha_t=1$.

\begin{algorithm}[ht]
  \caption{Fake Defensive Forecasting Algorithm with Discounting}
\label{alg:fullFDFD}
  \begin{algorithmic}[1]
   \STATE Initialize cumulative losses $\dL_0=0$, $\dL_0^k:=0$, $k=1,\ldots,K$.\\
          Set $\beta_1=1$, $B_0=0$.
   \FOR{$t=1,2,\dots$}
     \STATE Get discount $\alpha_{t-1}\in(0,1]$; 
            update $\beta_t=\beta_{t-1}/\alpha_{t-1}$, $B_t=B_{t-1}+\beta_t$.
     \STATE Compute $\eta_t=\sqrt{\beta_t/B_t}$.
     \STATE Get Experts' predictions
                             {$\gamma_t^k \in \Gamma$, $k=1,\ldots,K$}.
     \STATE \label{fullFDFDline:find}
            Find $\gamma\in\Gamma$ s.t. for all $\omega\in\Omega$
            \quad 
            $f_t(\gamma,\omega) \le C_t$,\\ 
            where $f_t$ and $C_t$
            are defined by~\eqref{eq:fakesuper} and~\eqref{eq:fakeconstant},
            respectively.
     \STATE \label{fullFDFDline:output} Output $\gamma_t := \gamma$.
     \STATE Get $\omega_t\in\Omega$.
     \STATE Update $\dL_t:=\alpha_{t-1}\dL_{t-1}+\lambda(\gamma_t,\omega_t)$.
     \STATE Update $\dL_t^k:=\alpha_{t-1}\dL_{t-1}^k+\lambda(\gamma_t^k,\omega_t)$,
                            $k=1,\ldots,K$.
   \ENDFOR.

  \end{algorithmic}
\end{algorithm}

Algorithm~\ref{alg:fullFDFD} in line~\ref{fullFDFDline:find}
uses the function
\begin{multline}\label{eq:fakesuper}
f_t(\gamma,\omega)=
\sum_{k=1}^K \frac{1}{K} \sum_{j=1}^\infty
\frac{c}{j^2}
 \exp\left( j\alpha_{t-1}\eta_t(\dL_{t-1}-\dL_{t-1}^k)
      -\frac{j^2\eta_t}{2\beta_t}\sum_{\tau=1}^{t-1}\beta_\tau\eta_\tau
    \right)
\\
\times
\exp\left( j\eta_t(\lambda(\gamma,\omega)-\lambda(\gamma_{t}^k,\omega))
      -\frac{j^2\eta_t^2}{2}
    \right) 
\end{multline}
and the constant
\begin{equation}\label{eq:fakeconstant}
C_t = 
\sum_{k=1}^K \frac{1}{K} \sum_{j=1}^\infty
\frac{c}{j^2}
\exp\left( j\alpha_{t-1}\eta_t(\dL_{t-1}-\dL_{t-1}^k)
    -\frac{j^2\eta_{t}}{2\beta_{t}}\sum_{\tau=1}^{t-1}\beta_\tau\eta_\tau
   \right)\,,
\end{equation}
where $1/c=\sum_{j=1}^\infty \frac{1}{j^2}$.

Algorithm~\ref{alg:fullFDFD} is more complicated
than Algorithm~\ref{alg:FDFD},
and the loss bound we get is weaker and holds for a narrower class of games.
However, this bound can be stated as 
a bound for \emph{$\epsilon$-quantile regret}
introduced in~\cite{CFH2009}.
Namely, let $\dL_t^\epsilon$
be any value such that for at least $\epsilon K$ Experts
their loss $\dL_t^k$ after step $t$
is not greater than $\dL_t^\epsilon$.
The $\epsilon$-quantile regret
is the difference between $\dL_t$ and $\dL_t^\epsilon$.
For $\epsilon=1/K$, we can choose 
$\dL_t^\epsilon=\min_k\dL_t^k\le \dL_t^k$ 
for all $k=1,\ldots,K$,
and thus a bound in terms of the $\epsilon$-quantile regret
implies a bound in terms of $\dL_t^k$.
The value $1/\epsilon$ plays the role of the ``effective'' number
of experts. 
Algorithm~\ref{alg:fullFDFD} guarantees a bound
in terms of $\dL_t^\epsilon$ for any $\epsilon>0$,
without the prior knowledge of $\epsilon$,
and in this sense the algorithm works for 
the unknown number of Experts
(see~\cite{CV2010} for a more detailed discussion).

For Algorithm~\ref{alg:fullFDFD} we need to restrict the class
of games we consider.
The game is called \emph{compact} if
the set
$\Lambda=
\{\lambda(\gamma,\cdot)\in\mathbb{R}^\Omega\givn \gamma\in\Gamma\}$
is compact in the standard topology of $\R^\Omega$.

\begin{theorem}\label{thm:convexsuperbound}
Suppose that $(\Omega,\Gamma,\lambda)$
is a non-empty convex compact game,
$\Omega$ is finite,
and $\lambda(\gamma,\omega)\in[0,1]$
for all $\gamma\in\Gamma$ and $\omega\in\Omega$.
In the game played according to Protocol~\ref{prot:GenDisc},
Learner has a strategy guaranteeing
that, for any $T$ and for any $\epsilon>0$, it holds
\begin{equation}\label{eq:sqrtsuperbound}
\dL_T \le \dL_T^\epsilon + 
  2\sqrt{\frac{B_T}{\beta_T}\ln\frac{1}{\epsilon}}+
  7\sqrt{\frac{B_T}{\beta_T}}\,,
\end{equation}
%
%
where $\beta_t=1/(\alpha_1\cdots\alpha_{t-1})$ and
$B_T=\sum_{t=1}^T \beta_t$.
\end{theorem}
\begin{proof}
The most difficult part of the proof is to show
that one can find $\gamma$ in line~\ref{fullFDFDline:find}
of Algorithm~\ref{alg:fullFDFD}.
We do not do this here, but refer to~\cite{CV2010};
the proof is literally the same as in~\cite[Theorem~9]{CV2010}
and is based on the supermartingale property of~$f_t$.
(The rest of the proof below also follows~\cite[Theorem~9]{CV2010};
the only difference is in the definition of $f_t$
and $C_t$.)

Let us check that $C_t\le 1$ for all $t$.
Clearly, $C_1=1$.
Assume that we have $C_t\le 1$.
This implies $f_t(\gamma_t,\omega_t)\le 1$ due to the choice of $\gamma_t$,
and thus $(f_t(\gamma_t,\omega_t))^{\alpha_t\eta_{t+1}/\eta_{t}}\le 1$.
Similarly to~\eqref{eq:alphaless1}, we have
$\alpha_t\eta_{t+1}/\eta_{t}\le 1$.
Since the function $x\mapsto x^\alpha$ is concave
for $\alpha\in(0,1]$, $x\ge 0$, we get
\begin{multline*}
1\ge \bigl(f_t(\gamma_t,\omega_t)\bigr)^{\alpha_t\eta_{t+1}/\eta_{t}}
\\
=
\left(\sum_{k=1}^K \frac{1}{K} \sum_{j=1}^\infty
\frac{c}{j^2}
 \exp\left( j\eta_t(\dL_{t}-\dL_{t}^k)
      -\frac{j^2\eta_t}{2\beta_t}\sum_{\tau=1}^{t}\beta_\tau\eta_\tau
    \right)
\right)^{\alpha_t\eta_{t+1}/\eta_{t}}
\\
\ge
\sum_{k=1}^K \frac{1}{K} \sum_{j=1}^\infty
\frac{c}{j^2}
 \left(\exp\left( j\eta_t(\dL_{t}-\dL_{t}^k)
      -\frac{j^2\eta_t}{2\beta_t}\sum_{\tau=1}^{t}\beta_\tau\eta_\tau
    \right)
 \right)^{\alpha_t\eta_{t+1}/\eta_{t}}
\\
=
\sum_{k=1}^K \frac{1}{K} \sum_{j=1}^\infty
\frac{c}{j^2}
 \exp\left( j\alpha_t\eta_{t+1}(\dL_{t}-\dL_{t}^k)
      -\frac{j^2\eta_{t+1}}{2\beta_{t+1}}\sum_{\tau=1}^{t}\beta_\tau\eta_\tau
    \right)
= C_{t+1}\,.
\end{multline*}

Thus, for each $t$ we have $f_t(\gamma_t,\omega_t)\le 1$,
that is,
$$
\sum_{k=1}^K \frac{1}{K} \sum_{j=1}^\infty
\frac{c}{j^2}
 \exp\left( j\eta_t(\dL_{t}-\dL_{t}^k)
      -\frac{j^2\eta_t}{2\beta_t}\sum_{\tau=1}^{t}\beta_\tau\eta_\tau
    \right)
\le 1\,.
$$
For any $\epsilon>0$, let us take any $\dL_T^\epsilon$
such that
for at least $\epsilon K$ Experts
their losses $\dL_T^k$ are smaller than or equal to $\dL_T^\epsilon$.
Then we have 
\begin{multline*}\allowdisplaybreaks
1\ge
\sum_{k=1}^K \frac{1}{K} \sum_{j=1}^\infty
\frac{c}{j^2}
 \exp\left( j\eta_t(\dL_{t}-\dL_{t}^k)
      -\frac{j^2\eta_t}{2\beta_t}\sum_{\tau=1}^{t}\beta_\tau\eta_\tau
    \right)
\\
\ge 
\epsilon
\sum_{j=1}^\infty
\frac{c}{j^2}
 \exp\left( j\eta_t(\dL_{t}-\dL_{t}^\epsilon)
      -\frac{j^2\eta_t}{2\beta_t}\sum_{\tau=1}^{t}\beta_\tau\eta_\tau
    \right)
\\
\ge
\frac{c\epsilon}{j^2}
 \exp\left( j\eta_t(\dL_{t}-\dL_{t}^\epsilon)
      -\frac{j^2\eta_t}{2\beta_t}\sum_{\tau=1}^{t}\beta_\tau\eta_\tau
    \right)
\end{multline*}
for any natural $j$.
Taking the logarithm and rearranging,
we get
$$
\dL_{t}\le \dL_{t}^\epsilon
   + \frac{j}{2\beta_t}\sum_{\tau=1}^{t}\beta_\tau\eta_\tau
   + \frac{1}{j\eta_t} \ln\frac{j^2}{c\epsilon}\,.
$$
Substituting $\eta_t=\sqrt{\beta_t/B_t}$
and using Lemma~\ref{lem:gen-sqrt-sum},
we get
$$
\dL_{t}\le \dL_{t}^\epsilon
   + \left(j+\frac{2}{j}\ln j + \frac{1}{j}\ln\frac{1}{\epsilon}
                              + \frac{1}{j}\ln\frac{1}{c}
     \right)
     \sqrt{\frac{B_t}{\beta_t}}\,.
$$
Letting $j=\left\lceil\sqrt{\ln(1/\epsilon)}\right\rceil+1$
and using the estimates $j\le \sqrt{\ln(1/\epsilon)}+2$,
$(\ln j)/j\le 2$, 
$(\ln(1/\epsilon))/j\le \sqrt{\ln(1/\epsilon)}$,
$1/j\le 1$,
and $\ln(1/c)=\ln(\pi^2/6)\le 1$,
we obtain the final bound.
\EP 

\fi

\section{Regression with Discounted Loss}
\label{sec:regression}

In this section we consider a task of regression,
where Learner must predict ``labels'' $y_t\in\R$
for input instances $x_t\in\bX\subseteq\R^n$.
The predictions proceed according to Protocol~\ref{prot:COP}.
\begin{protocol}[h]
  \caption{Competitive online regression}
  \label{prot:COP}
  \begin{algorithmic}
    \FOR{$t=1,2,\dots$}
      \STATE Reality announces $x_t\in\bX$.
      \STATE Learner announces $\gamma_t\in\Gamma$.
      \STATE Reality announces $y_t\in\Omega$.
    \ENDFOR
  \end{algorithmic}
\end{protocol}
This task can be embedded into prediction with expert advice
if Learner competes with all functions $x\to y$
from some large class serving as a pool of
(imaginary) Experts.

\subsection{The Framework and Linear Functions as Experts}

Let the input space be $\bX\subseteq\R^n$,
the set of predictions be $\Gamma = \R$,
and
the set of outcomes be $\Omega = [Y_1,Y_2]$.
In this section we consider the square loss
$\lambda^\mathrm{sq}(\gamma,y) = (\gamma-y)^2$.
Learner competes with a pool of experts
$\Theta=\R^n$ (treated as linear functionals on $\R^n$).
Each individual expert is denoted by~$\theta\in\Theta$
and predicts $\theta'x_t$ at step $t$.

Let us take any distribution over the experts $P(d\theta)$.
It is known from \cite{VovkAS} that~\eqref{eq:realiz} holds for the square loss
with $c=1$, $\eta = \frac{2}{(Y_2-Y_1)^2}$:
\begin{equation}\label{eq:linrealiz}
\exists \gamma\in\Gamma \:\forall y\in\Omega=[Y_1,Y_2]\quad
(\gamma - y)^2\le
-\frac{1}{\eta}
  \ln\left(\int_\Theta \e^{-\eta (\theta'x_t - y)^2}P(d\theta)\right).
\end{equation}

Denote by $X$ the matrix of size $T\times n$
consisting of the rows of the input vectors $x_1',\ldots,x_T'$.
Let also $W_T = \diag(\beta_1/\beta_T,\beta_2/\beta_T,\ldots,\beta_T/\beta_T)$, i.e.,
$W_T$ is a diagonal matrix $T \times T$.
In a manner similar to~\cite{VovkCOS},
we prove the following upper bound for Learner's loss.
\begin{theorem}\label{thm:linbound}
For any $a > 0$,
there exists a prediction strategy for Learner in Protocol~\ref{prot:COP} 
achieving, for every $T$
and  for any linear predictor $\theta \in \R^n$,
  \begin{multline}\label{eq:linbound}
  \sum_{t=1}^T \frac{\beta_t}{\beta_T} (\gamma_t-y_t)^2
  \le
  \sum_{t=1}^T \frac{\beta_t}{\beta_T} (\theta'x_t-y_t)^2 
\\
  + a\|\theta\|^2
  + \frac{(Y_2-Y_1)^2}{4}\ln\det\left(\frac{X'W_TX}{a} + I\right)\,.
  \end{multline}
If, in addition, $\|x_t\|_\infty \le Z$ for all $t$, then
  \begin{multline}\label{eq:linboundT}
  \sum_{t=1}^T \frac{\beta_t}{\beta_T} (\gamma_t-y_t)^2
  \le
  \sum_{t=1}^T \frac{\beta_t}{\beta_T} (\theta'x_t-y_t)^2 
\\
  + a\|\theta\|^2
  + \frac{n(Y_2-Y_1)^2}{4}
     \ln\left(\frac{Z^2}{a}\frac{\sum_{t=1}^T \beta_t}{\beta_T} + 1
                   \right).
  \end{multline}
\end{theorem}

In the undiscounted case ($\alpha_t=1$ for all $t$),
the bounds in the theorem coincide with the bounds
for the Aggregating Algorithm for Regression~\cite[Theorem~1]{VovkCOS}
with $Y_2 = Y$ and $Y_1 = -Y$,
since, as remarked after Theorem~\ref{thm:convexbound},
$\beta_t=1$ and
$\left(\sum_{t=1}^T \beta_t\right)/\beta_T = T$
in the undiscounted case. 
Recall also that in the case of the exponential discounting
($\alpha_t = \alpha \in(0,1)$)
we have $\beta_t=\alpha^{-t+1}$ and
$\left(\sum_{t=1}^T \beta_t\right)/\beta_T = (1-\alpha^{T-1})/(1-\alpha)
\le 1/(1-\alpha)$.
\ifARXIV
Thus, for the exponential discounting bound~\eqref{eq:linboundT}
becomes
\begin{multline}\label{eq:linboundconst}
  \sum_{t=1}^T \alpha^{T-t} (\gamma_t-y_t)^2
  \le
  \sum_{t=1}^T \alpha^{T-t} (\theta'x_t-y_t)^2 
\\
  + a\|\theta\|^2
  + \frac{n(Y_2-Y_1)^2}{4}
     \ln\left(\frac{Z^2(1-\alpha^{T-1})}{a(1-\alpha)} + 1\right)\,.
\end{multline}
\fi

\subsection{Functions from an RKHS as Experts}
In this section we apply the kernel trick
to the linear method
to compete with wider sets of experts.
Each expert $f\in\hs{F}$ predicts $f(x_t)$.
Here $\hs{F}$ is a reproducing kernel Hilbert space (RKHS)
with a positive definite kernel $k \colon \bX\times\bX \to \R$.
For the definition of RKHS and
its connection to kernels see \cite{Scholkopf2002}.
Each kernel defines a unique RKHS.
We use the notation $\mat{K}_T = \{k(x_i,x_j)\}_{i,j=1,\ldots,T}$
for the kernel matrix
for the input vectors at step~$T$.
In a manner similar to~\cite{KAAR},
we prove the following upper bound
on the discounted square loss of Learner.
\begin{theorem}\label{thm:hilbound}
  For any $a > 0$,
  there exists a strategy for Learner
  in Protocol~\ref{prot:COP}
  achieving,
  for every positive integer $T$
  and any predictor $f \in \hs{F}$,
  \begin{multline}\label{eq:hilbound}
  \sum_{t=1}^T \frac{\beta_t}{\beta_T} (\gamma_t-y_t)^2
  \le
  \sum_{t=1}^T \frac{\beta_t}{\beta_T} (f(x_t)-y_t)^2 
\\
  + a\|f\|^2
  + \frac{(Y_2-Y_1)^2}{4}
      \ln\det\left(\frac{\sqrt{W_T}\mat{K}_T\sqrt{W_T}}{a} + I
                      \right)\,.
  \end{multline}
\end{theorem}

\begin{corollary}\label{cor:hilboundconst}
Assume that 
$c^2_\hs{F} = \sup_{x \in \bX} k(x,x) < \infty$
for the RKHS $\hs{F}$.
Under the conditions of Theorem~\ref{thm:hilbound},
given in advance any constant $\mathcal{T}$ such that
$\left(\sum_{t=1}^T \beta_t\right)/\beta_T \le \mathcal{T}$,
one can choose parameter $a$ such that
the strategy in Theorem~\ref{thm:hilbound}
achieves for any $f \in \hs{F}$
 \begin{multline}\label{eq:hilboundconst}
    \sum_{t=1}^T \frac{\beta_t}{\beta_T} (\gamma_t-y_t)^2
  \le
  \sum_{t=1}^T \frac{\beta_t}{\beta_T} (f(x_t)-y_t)^2
  +
\left(\frac{(Y_2-Y_1)^2}{4}+\|f\|^2\right)c_{\hs{F}}\sqrt{\mathcal{T}}\,.
  \end{multline}
where $c^2_\hs{F} = \sup_{x \in \bX} k(x,x) < \infty$
characterizes the RKHS $\hs{F}$.
\end{corollary}
\BP
The determinant of a symmetric positive definite matrix
is upper bounded by the product of its diagonal elements
(see Chapter 2, Theorem 7 in \cite{Beckenbach1961}),
and thus we have
  \begin{multline*}
  \ln\det \left(I + \frac{\sqrt{W_T}\mat{K}_T\sqrt{W_T}}{a} \right)
  \le
  T \ln \left(1+\frac{c^2_{\hs{F}}\left(\prod_{t=1}^T \frac{\beta_t}{\beta_T}\right)^{1/T}}{a}\right)
  \\
\le
  T \frac{c^2_{\hs{F}}}{a}\left(\prod_{t=1}^T \frac{\beta_t}{\beta_T}\right)^{1/T}
\le
  T \frac{c^2_{\hs{F}}}{a\beta_T}\frac{\sum_{t=1}^T \beta_t}{T}
\le
\frac{c^2_{\hs{F}}\mathcal{T}}{a}
  \end{multline*}
(we use $\ln(1+x)\le x$ and the inequality between the geometric
and arithmetic means).
Choosing $a=c_{\hs{F}}\sqrt{\mathcal{T}}$,
we get bound~\eqref{eq:hilboundconst} from~\eqref{eq:hilbound}.
\EP
Recall again that
$\left(\sum_{t=1}^T \beta_t\right)/\beta_T = (1-\alpha^{T-1})/(1-\alpha)
\le 1/(1-\alpha)$ in the case of the exponential discounting
($\alpha_t = \alpha \in(0,1)$),
and we can take $\mathcal{T}=1/(1-\alpha)$.

In the undiscounted case ($\alpha_t=1$),
we have $\left(\sum_{t=1}^T \beta_t\right)/\beta_T=T$,
so we need to know the number of steps in advance.
Then, bound~\eqref{eq:hilboundconst} matches the bound
obtained in~\cite[the displayed formula after~(33)]{VovkRKHSarXiv}.
If we do not know an upper bound $\mathcal{T}$ in advance,
it is still possible to achieve a bound similar to~\eqref{eq:hilboundconst}
using the Aggregating Algorithm with Discounting
to merge Learner's strategies from Theorem~\ref{thm:hilbound}
with different values of parameter $a$,
in the same manner as in~\cite[Theorem~3]{VovkRKHSarXiv}.

\ifARXIV
\begin{corollary}\label{cor:hilboundmixed}
Assume that 
$c^2_\hs{F} = \sup_{x \in \bX} k(x,x) < \infty$
for the RKHS $\hs{F}$.
Under the conditions of Theorem~\ref{thm:hilbound},
  there exists a strategy for Learner
  in Protocol~\ref{prot:COP}
  achieving,
  for every positive integer $T$
  and any predictor $f \in \hs{F}$,
 \begin{multline}\label{eq:hilboundmixed}
    \sum_{t=1}^T \frac{\beta_t}{\beta_T} (\gamma_t-y_t)^2
  \le
  \sum_{t=1}^T \frac{\beta_t}{\beta_T} (f(x_t)-y_t)^2
  +
c_{\hs{F}}\|f\|(Y_2-Y_1)\sqrt{\frac{\sum_{t=1}^T \beta_t}{\beta_T}}
\\
  +\frac{(Y_2-Y_1)^2}{2}\ln\frac{\sum_{t=1}^T \beta_t}{\beta_T}
  + \|f\|^2
  + (Y_2-Y_1)^2\ln\left(\frac{c_{\hs{F}}(Y_2-Y_1)}{\|f\|}+2 
                  \right)\,.
  \end{multline}
\end{corollary}
\BP
Let us take the strategies from Theorem~\ref{thm:hilbound}
for $a=1,2,3,\ldots$
and provide them as Experts 
to the Aggregating Algorithm with Discounting,
with the square loss function,
$\eta=2/(Y_2-Y_1)^2$ and initial Experts' weights proprotional
to $1/a^2$.
Then Theorem~\ref{thm:AAD} 
(extended as described in Remark at the end of Section~\ref{sec:linear})
guarantees that the extra loss of the aggregated
strategy (compared to the strategy from Theorem~\ref{thm:hilbound}
with parameter $a$) is not greater than 
$\frac{(Y_2-Y_1)^2}{2}\ln\frac{a^2}{c}$, where $c=\sum_{k=1}^K 1/k^2$.
On the other hand,
for the strategy from Theorem~\ref{thm:hilbound}
with parameter $a$ similarly to the proof of Corollary~\ref{cor:hilboundconst}
we get
$$
  \sum_{t=1}^T \frac{\beta_t}{\beta_T} (\gamma_t-y_t)^2
  \le
  \sum_{t=1}^T \frac{\beta_t}{\beta_T} (f(x_t)-y_t)^2 
  + a\|f\|^2
  + \frac{c_{\hs{F}}^2(Y_2-Y_1)^2}{4a}\frac{\sum_{t=1}^T \beta_t}{\beta_T}\,.
$$
Adding $\frac{(Y_2-Y_1)^2}{2}\ln\frac{a^2}{c}$ to the right-hand side
and choosing 
$$
a=\left\lceil \frac{c_{\hs{F}}(Y_2-Y_1)}{2\|f\|}
              \sqrt{\frac{\sum_{t=1}^T \beta_t}{\beta_T}}\,
   \right\rceil\,,
$$
we get the statement after simple estimations.
\EP
\fi

\subsection{Proofs of Theorems~\ref{thm:linbound} and~\ref{thm:hilbound}}
Let us begin with several technical lemmas from linear algebra.
\ifCONF
For complete proofs of them see~\cite[Appendix~A.1]{CZ2010}.
\fi
\ifARXIV
The proofs of some of these lemmas are moved to
Appendix~\ref{append:technical}.
\fi

\begin{lemma}\label{lem:integraleval}
Let $A$ be a symmetric positive definite matrix of size~$n \times n$.
Let $\theta, b \in \R^n$, $c$ be a real number,
and $Q(\theta)=\theta'A\theta + b'\theta + c$.
Then
  \begin{equation*}
    \int_{\R^n} e^{-Q(\theta)} d\theta = e^{-Q_0} \frac{\pi^{n/2}}{\sqrt{\det A}},
  \end{equation*}
  where $Q_0 = \min_{\theta \in \R^n} Q(\theta)$.
\end{lemma}
\noindent
The proof of this lemma can be found
in~\cite[Theorem~15.12.1]{Harville1997}.

\begin{lemma}\label{lem:Frepres}
Let $A$ be a symmetric positive definite  matrix of size~$n \times n$.
Let $b,z \in \R^n$, and
  \begin{equation*}
    F(A,b,z) = \min_{\theta \in \R^n}(\theta' A \theta + b'\theta + z'\theta)
    - \min_{\theta \in \R^n}(\theta' A \theta + b'\theta - z'\theta)\,.
  \end{equation*}
Then $F(A,b,z) = -b'A^{-1}z$.
\end{lemma}

\begin{lemma}\label{lem:ratiointegr}
Let $A$ be a symmetric positive definite matrix of size~$n \times n$.
Let $\theta,b_1,b_2 \in \R^n$, $c_1, c_2$ be real numbers,
and
$Q_1(\theta)=\theta'A\theta + b_1'\theta + c_1$,
$Q_2(\theta)=\theta'A\theta + b_2'\theta + c_2$.
Then
  \begin{equation*}
  \frac{\int_{\R^n} e^{-Q_1(\theta)} d\theta}
       {\int_{\R^n} e^{-Q_2(\theta)} d\theta}
       = e^{c_2-c_1 - \frac{1}{4}(b_2+b_1)'A^{-1}(b_2-b_1)}\,.
  \end{equation*}
\end{lemma}

The previous three lemmas
were implicitly used in \cite{VovkCOS} to derive a bound
on the cumulative undiscounted square loss
of the algorithm competing with linear experts.

\begin{lemma}\label{lem:matrixequal}
For any matrix $B$ of size~$n\times m$,
any matrix $C$ of size~$m\times n$,
and any real number $a$
such that the matrices $aI_m+CB$ and $aI_n+BC$ are nonsingular,
it holds
\begin{equation}\label{eq:matrixequal}
  B(aI_m+CB)^{-1}=(aI_n+BC)^{-1}B\,,
\end{equation}
where $I_n,I_m$ are the unit matrices of sizes~$n\times n$ and~$m\times m$,
respectively.
\end{lemma}
\BP
Note that this is equivalent to $(aI_n+BC)B=B(aI_m+CB)$.
\EP

\begin{lemma}\label{lem:matdetiden}
For matrix $B$ of size~$n\times m$,
any matrix $C$ of size~$m\times n$,
and any real number $a$, it holds
  \begin{equation}\label{eq:matdetiden}
  \det (aI_n + BC)=\det (aI_m + CB)\,,
  \end{equation}
where $I_n,I_m$ are the unit matrices of sizes~$n\times n$ and~$m\times m$,
respectively.
\end{lemma}

\subsubsection{Proof of Theorem~\ref{thm:linbound}.}
We take the Gaussian initial distribution over the experts
with a parameter $a>0$:
\begin{equation*}
P_0(d\theta) = \left(\frac{a\eta}{\pi}\right)^{n/2} e^{-a\eta\|\theta\|^2}d\theta.
\end{equation*}
and use ``Algorithm~\ref{alg:AAD} with infinitely many Experts''.
Repeating the derivations from
\ifCONF the proof of Theorem~\ref{thm:AAD}, \fi
\ifARXIV Subsection~\ref{ssec:AAproof}, \fi
we obtain the following analogue of~\eqref{eq:AADFbound}:
\begin{equation*}
\left(\frac{a\eta}{\pi}\right)^{n/2}
\int_\Theta \e^{\eta\left(\sum_{t=1}^T \frac{\beta_t}{\beta_T} (\gamma_t-y_t)^2
-
\sum_{t=1}^T \frac{\beta_t}{\beta_T} (\theta'x_t-y_t)^2\right)} e^{-a\eta\|\theta\|^2}d\theta \le 1.
\end{equation*}

The simple equality
\begin{equation}\label{eq:exlosstrans}
\sum_{t=1}^T \frac{\beta_t}{\beta_T} (\theta'x_t-y_t)^2 + a\|\theta\|^2
=
  \theta'(aI + X'W_TX)\theta
 - 2\sum_{t=1}^T \frac{\beta_t}{\beta_T} y_t\theta'x_t
  + \sum_{t=1}^T \frac{\beta_t}{\beta_T} y_t^2
\end{equation}
shows that
the integral can be evaluated with the help of Lemma~\ref{lem:integraleval}:
\begin{multline*}
\left(\frac{a\eta}{\pi}\right)^{n/2}
\int_\Theta e^{-\eta \left(\sum_{t=1}^T \frac{\beta_t}{\beta_T} (\theta'x_t-y_t)^2 + a\|\theta\|^2 \right)} d\theta
\\ =
\left(\frac{a}{\pi}\right)^{n/2}
e^{-\eta \min_{\theta} \left(\sum_{t=1}^T \frac{\beta_t}{\beta_T} (\theta'x_t-y_t)^2 + a\|\theta\|^2 \right)} 
      \frac{\pi^{n/2}}{\sqrt{\det(aI + X'W_TX)}}.
\end{multline*}
We take the natural logarithms of both parts of the bound and using the value $\eta = \frac{2}{(Y_2-Y_1)^2}$
obtain~\eqref{eq:linbound}.
The determinant of a symmetric positive definite matrix
is upper bounded by the product of its diagonal elements
(see Chapter 2, Theorem 7 in \cite{Beckenbach1961}):
\begin{equation*}
\det\left(\frac{X'W_TX}{a} + I\right)
  \le \left(\frac{Z^2\sum_{t=1}^T \beta_t}{a\beta_T} + 1\right)^n,
\end{equation*}
and thus we obtain~\eqref{eq:linboundT}.

\subsubsection{Proof of Theorem~\ref{thm:hilbound}.}
  We must prove that for each $T$
  and each sequence $(x_1,y_1,\ldots,x_T,y_T)\in(\bX\times\R)^T$
  the guarantee~\eqref{eq:hilbound} is satisfied.
  Fix $T$ and $(x_1,y_1,\ldots,x_T,y_T)$.
  Fix an isomorphism between the linear span of $k_{x_1},\ldots,k_{x_T}$
  obtained for the Riesz Representation theorem
  and $\R^{\tilde T}$,
  where $\tilde T\le T$ is the dimension of the linear span of $k_{x_1},\ldots,k_{x_T}$.
  Let $\tilde x_1,\ldots,\tilde x_T\in\R^{\tilde T}$ be the images of $k_{x_1},\ldots,k_{x_T}$,
  respectively,
  under this isomorphism.
  We have then $k(\cdot,x_i) = \langle \cdot,\tilde x_i \rangle$ for any $x_i$.

  We apply the strategy from  Theorem~\ref{thm:linbound}
  to $\tilde x_1,\ldots,\tilde x_T$.
  The predictions of the strategies are the same due to 
  Proposition~\ref{prop:hilpred} below.
  Any expert $\theta\in\R^{\tilde T}$ in bound~\eqref{eq:linbound} 
  can be represented as
  \begin{equation*}
  \theta = \sum_{i=1}^T c_i \tilde x_i = \sum_{i=1}^T c_i k(\cdot,x_i)
  \end{equation*}
  for some $c_i\in\R$.
  Thus the experts' predictions are 
  $\theta' \tilde x_t = \sum_{i=1}^T c_i k(x_t,x_i)$,
  and the norm is 
  $\|\theta\|^2 = \sum_{i,j=1}^T c_i c_j k(x_i,x_j)$.

  Denote by $\tilde X$ the $T\times \tilde T$
  matrix consisting of the rows of the vectors $\tilde x_1',\ldots,\tilde x_T'$.
  From Lemma~\ref{lem:matdetiden} we have
  \begin{equation*}
  \det\left(\frac{\tilde X'W_T\tilde X}{a} + I\right)
  =
  \det\left(\frac{\sqrt{W_T}\tilde X \tilde X'\sqrt{W_T}}{a} + I\right).
  \end{equation*}
  Thus using $\mat{K}_T = \tilde X \tilde X'$ we obtain the upper bound
  \begin{multline*}
  \sum_{t=1}^T \frac{\beta_t}{\beta_T} (\gamma_t-y_t)^2
  \le
  \sum_{t=1}^T \frac{\beta_t}{\beta_T} \left(\sum_{i=1}^T c_i k(x_t,x_i)-y_t\right)^2 
\\
  + a\sum_{i,j=1}^T c_i c_j k(x_i,x_j)
  + \frac{(Y_2-Y_1)^2}{4}\ln\det\left(\frac{\sqrt{W_T}\mat{K}_T\sqrt{W_T}}{a} + I\right)
  \end{multline*}
  for any $c_i\in\R$, $i=1,\ldots,T$.
  By the Representer theorem (see Theorem 4.2 in \cite{Scholkopf2002})
  the minimum of $\sum_{t=1}^T \frac{\beta_t}{\beta_T} (f(x_t)-y_t)^2 + a\|f\|^2$ over all $f\in\hs{F}$
  is achieved
  on one of the linear combinations from the bound obtained above.
  This concludes the proof.

\subsection{Regression Algorithms}\label{sec:algreg}
In this subsection we derive explicit form of
the prediction strategies for Learner
used in Theorems~\ref{thm:linbound}
and~\ref{thm:hilbound}.

\subsubsection{Strategy for Theorem~\ref{thm:linbound}.}
In~\cite{VovkCOS} Vovk suggests for the square loss
the following substitution function
satisfying~\eqref{eq:subst}:
\begin{equation}\label{eq:substsquare}
\gamma_T = \frac{Y_2+Y_1}{2} - \frac{g_T(Y_2)-g_T(Y_1)}{2(Y_2-Y_1)}.
\end{equation}
It allows us to calculate $g_T$ with unnormalized weights:
\begin{equation*}
g_T(y)=
    -\frac{1}{\eta}\left(\ln\int_{\Theta} e^{-\eta\left(\theta'A_T\theta -
                             2\theta'\left(b_{T-1} + yx_T\right)
                             +
                             \left(\sum_{t=1}^{T-1} \frac{\beta_t}{\beta_T} y_t^2 + y^2\right)\right)}d\theta\right)
\end{equation*}
for any $y\in\Omega=[Y_1,Y_2]$ 
(here we use the expansion~\eqref{eq:exlosstrans}),
where
\begin{equation*}
A_T = aI + \sum_{t=1}^{T-1} \frac{\beta_t}{\beta_T}x_t x_t' + x_T x_T' = aI + X'W_TX,
\end{equation*}
and $b_{T-1} = \sum_{t=1}^{T-1} \frac{\beta_t}{\beta_T} y_t x_t$.
The direct calculation of $g_T$
is inefficient: it requires numerical integration.
Instead, we notice that
\begin{multline}
\gamma_T = \frac{Y_2+Y_1}{2} - \frac{g_T(Y_2)-g_T(Y_1)}{2(Y_2-Y_1)}
  \\
         = \frac{Y_2+Y_1}{2}
          - \frac{1}{2(Y_2-Y_1)\eta}
            \ln \frac{\int_{\Theta} e^{-\eta\left(\theta'A_T\theta -
                             2\theta'\left(b_{T-1} + Y_1x_T\right)
                             +
                             \left(\sum_{t=1}^{T-1} \frac{\beta_t}{\beta_T} y_t^2 + Y_1^2\right)\right)}d\theta}
                             {\int_{\Theta} e^{-\eta\left(\theta'A_T\theta -
                             2\theta'\left(b_{T-1} + Y_2x_T\right)
                             +
                             \left(\sum_{t=1}^{T-1} \frac{\beta_t}{\beta_T} y_t^2 + Y_2^2\right)\right)}d\theta}
  \\[0.5ex]
       = \frac{Y_2+Y_1}{2}
         - \frac{1}{2(Y_2-Y_1)\eta}
           \ln e^{\eta\left(Y_2^2-Y_1^2
           - \left(b_{T-1}+\left(\frac{Y_2+Y_1}{2}\right)x_T\right)'A_T^{-1}
           \left(\frac{Y_2-Y_1}{2}x_T\right)\right)}
  \\
       = \left(b_{T-1}+\left(\frac{Y_2+Y_1}{2}\right)x_T\right)'A_T^{-1} x_T\,,
\label{eq:linpred}
\end{multline}
where the third equality follows from Lemma~\ref{lem:ratiointegr}.

The strategy which predicts according to~\eqref{eq:linpred} 
requires $O(n^3)$ operations per step.
The most time-consuming operation is the inverse of the matrix $A_T$.
\ifARXIV
Note that for the undiscounted case the inverse could be computed
incrementally using the Sherman-Morrison formula,
which leads to $O(n^2)$ operations per step.
\fi

\subsubsection{Strategy for Theorem~\ref{thm:hilbound}.}
We use following notation.
Let
\begin{equation}\label{eq:hilnotation}
\begin{array}{lcl}
\vect{k}_T  & \text{be} & \text{the last column of the matrix } \mat{K}_T,\vect{k}_T = \{k(x_i,x_T)\}_{i=1}^T,\\
\vect{Y}_T  & \text{be} & \text{the column vector of the outcomes } \vect{Y}_T = (y_1,\ldots,y_T)'.
\end{array}
\end{equation}
When we write $\vect{Z} = (\vect{V};\vect{Y})$ or $\vect{Z} = (\vect{V}';\vect{Y}')'$ we mean that
the column vector $\vect{Z}$ is obtained by
concatenating two column vectors $\vect{V},\vect{Y}$ vertically or
$\vect{V}',\vect{Y}'$ horizontally.

As it is clear from the proof of Theorem~\ref{thm:hilbound},
we need to prove that the strategy for this theorem
is the same as the strategy for Theorem~\ref{thm:linbound}
in the case when the kernel is the scalar product.

\begin{proposition}\label{prop:hilpred}
 The predictions~\eqref{eq:linpred} can be represented as
 \begin{equation}\label{eq:hilpred}
 \gamma_T = \left( \vect{Y}_{T-1}; \frac{Y_2+Y_1}{2} \right)'\sqrt{W_T}
 \left(aI + \sqrt{W_T}\mat{K}_T\sqrt{W_T}\right)^{-1} \sqrt{W_T}\vect{k}_T
 \end{equation}
 for the scalar product kernel $k(x,y) = \langle x, y \rangle$, the unit $T \times T$ matrix $I$, and $a>0$.
\end{proposition}
\BP
  For the scalar product kernel we have we have $\mat{K}_T = X'X$
  and $\sqrt{W_T}\vect{k}_T = \sqrt{W_T} X x_T$.
  By Lemma~\ref{lem:matrixequal} we obtain
  \begin{equation*}
  \left(aI + \sqrt{W_T}XX'\sqrt{W_T}\right)^{-1} \sqrt{W_T} X x_T
  =
  \sqrt{W_T} X \bigl(aI + X'W_TX\bigr)^{-1}x_T\,.
  \end{equation*}
  It is easy to see that
  \begin{equation*}
  \left( \vect{Y}_{T-1}; \frac{Y_2+Y_1}{2}\right)'W_T X
  =
  \left(\sum_{t=1}^{T-1} \frac{\beta_t}{\beta_T} y_t x_t+\left(\frac{Y_2+Y_1}{2}\right)x_T\right)'
  \end{equation*}
  and
  \begin{equation*}
   X'W_TX = \sum_{t=1}^{T-1} \frac{\beta_t}{\beta_T} x_t x_t' + x_T x_T'\,.
  \end{equation*}
  Thus we obtain the formula~\eqref{eq:linpred} from~\eqref{eq:hilpred}.
\EP

\subsubsection*{Acknowledgements}
We are grateful to Yura Kalnishkan and Volodya Vovk
for numerous illuminating discussions.
This work was supported by EPSRC
(grant EP/F002998/1).

\ifARXIV
\appendix

\section{Appendix}

\subsection{Proofs of Technical Lemmas}
\label{append:technical}

\begin{proof}[Proof of Lemma~\ref{lem:gen-sqrt-sum}]
For $T=1$ the inequality is trivial.
Assume it for $T-1$. Then
\begin{multline*}
\frac{1}{\beta_T}\sum_{t=1}^T \beta_t\sqrt{\frac{\beta_t}{B_t}}
=
\frac{\beta_{T-1}}{\beta_T}\left(
           \frac{1}{\beta_{T-1}}\sum_{t=1}^{T-1}\beta_t\sqrt{\frac{\beta_t}{B_t}}
                           \right)
+\sqrt{\frac{\beta_T}{B_T}}
\\
\le
2\frac{\beta_{T-1}}{\beta_T}\sqrt{\frac{B_{T-1}}{\beta_{T-1}}}
+\sqrt{\frac{\beta_T}{B_T}}
=2\sqrt{\frac{\beta_{T-1}}{\beta_T}}\sqrt{\frac{B_{T-1}}{\beta_T}}
+\sqrt{\frac{\beta_T}{B_T}}
\\
\le
2\sqrt{\frac{B_{T-1}}{\beta_T}}
+\sqrt{\frac{\beta_T}{B_{T-1}+\beta_T}}
\le
2\sqrt{\frac{B_{T-1}+\beta_T}{\beta_T}}
=
2\sqrt{\frac{B_T}{\beta_T}}
\,.
\end{multline*}
The first inequality is by the induction assumption,
and the second inequality holds since $\beta_{T-1}\le\beta_T$.
The last inequality is
$2\sqrt{x}/\sqrt{y} + \sqrt{y}/\sqrt{x+y} \le
2\sqrt{x+y}/\sqrt{y}$, which holds for any positive $x$ and $y$.
(Indeed, it is equivalent to
$2\sqrt{x}\sqrt{x+y} + y \le {2(x+y)}$ and
$2\sqrt{x}\sqrt{x+y} \le x+y + x$.)
\EP

\begin{proof}[Proof of Lemma~\ref{lem:Frepres}]
  This lemma is proven by taking the derivative
  of the quadratic forms in $F$ by $\theta$ and calculating the minimum:
  $\min_{\theta \in \R^n}(\theta' A \theta + c'\theta) = -\frac{(A^{-1}c)'}{4}c$
  for any~$c \in \R^n$ (see Theorem 19.1.1 in \cite{Harville1997}).
\EP

\begin{proof}[Proof of Lemma~\ref{lem:ratiointegr}]
  After evaluating each of the integrals using Lemma~\ref{lem:integraleval}
  the ratio is represented as follows:
  \begin{equation*}
  \frac{\int_{\R^n} e^{-Q_1(\theta)} d\theta}
       {\int_{\R^n} e^{-Q_2(\theta)} d\theta}
       = e^{\min_{\theta \in \R^n} Q_2(\theta) - \min_{\theta \in \R^n} Q_1(\theta)} \enspace.
  \end{equation*}
  The difference of minimums can be calculated using Lemma~\ref{lem:Frepres}
  with~$b = \frac{b_2+b_1}{2}$ and~$z = \frac{b_2-b_1}{2}$:
  \begin{equation*}
  \min_{\theta \in \R^n} Q_2(\theta) - \min_{\theta \in \R^n} Q_1(\theta) =
  c_2-c_1 - \frac{1}{4}(b_2+b_1)'A^{-1}(b_2-b_1)\enspace.
  \end{equation*}
\EP

\begin{proof}[Proof of Lemma~\ref{lem:matdetiden}]
Consider the product of block matrices:
  \begin{equation*}
  \begin{pmatrix}
    I_n & B\\
    0 & I_m\\
  \end{pmatrix}
  \begin{pmatrix}
    aI_n+BC & 0\\
    -C & aI_m\\
  \end{pmatrix}=
  \begin{pmatrix}
    aI_n& aB\\
    -C & aI_m\\
  \end{pmatrix}
  =
  \begin{pmatrix}
    aI_n & 0\\
    -C & aI_m+CB\\
  \end{pmatrix}
  \begin{pmatrix}
    I_n & B\\
    0 & I_m\\
  \end{pmatrix}
  \end{equation*}
Taking the determinant of both sides, and
using formulas for the determinant of a block matrix,
we get the statement of the lemma.
\EP

\subsection{An Alternative Derivation of Regression Algorithms
Using Defensive Forecasting}
\label{append:DF}
In this section we derive the upper bound and the algorithms
using a different technique,
the defensive forecasting~\cite{Chernov2010}.

\subsubsection{Description of the Proof Technique}
We denote the predictions of any expert $\theta$
(from a finite set or following strategies from Section~\ref{sec:regression})
by $\xi_t^\theta$.
For each step $T$ and each expert $\theta$ we define the function
\begin{equation}\label{eq:Qtheta}
\begin{split}
&Q^\theta_t: \Gamma \times \Omega \to [0,\infty) \\
&Q^\theta_t(\gamma,y) := e^{\eta\left( \lambda(\gamma,y)
                            - \lambda(\xi_t^\theta,y)\right)}.
\end{split}
\end{equation}
We also define the mixture function
\begin{equation*}
Q_T := \int_\Theta \prod_{t=1}^{T-1} \left(Q^\theta_t\right)^{\prod_{i=t}^{T-1}\alpha_i} Q^\theta_T P_0(d\theta)
\end{equation*}
with some initial weights distribution~$P_0(d\theta)$ on the experts.
Here~$\eta$ is a learning rate coefficient;
it will be defined later in the section.
We define the correspondence
\begin{equation}\label{eq:pigamma}
\gamma^p = p (Y_2 - Y_1) + Y_1, \quad p \in [0,1],
\end{equation}
between $[0,1]$ and Learner's predictions $\gamma^p \in \Gamma$.

Let us introduce the notion of a defensive property.
We use the notation $\delta\Omega :=  \{Y_1,Y_2\}$.
Assume that there is a fixed bijection between
the space $\PPP(\delta\Omega)$ of all probability measures on $\delta\Omega$
and the set $[0,1]$. 
Each $p^\pi \in[0,1]$ corresponds to some unique  $\pi \in\PPP(\delta\Omega)$.

\begin{definition}
\rm{A sequence $R$ of functions $R_1,R_2,\ldots$ such that
$R_t: \Gamma \times \Omega \to(-\infty,\infty]$
is said to have \defin{the defensive property}
if, for any $T$
and any $\pi_T \in \PPP(\delta\Omega)$,
it holds that
\begin{equation}\label{eq:fin_DFsupersdefin}
  \EXP_{\pi_T} R_T(\gamma^{p^{\pi_T}},y)  \le 1,
\end{equation}
where $\EXP_\pi$ is
the expectation
with respect to a measure $\pi$.}
\end{definition}

A sequence $R$ is called \defin{forecast-continuous} if,
for all $T$
and all $y\in\Omega$,
all the functions $R_T(\gamma,y)$
are continuous in $\gamma$.

We now prove that $Q^\theta_t$ has the defensive property.
\begin{lemma}\label{lem:defprop}
  For~$\eta\in\left(0,\frac{2}{(Y_2-Y_1)^2}\right]$
  \begin{equation*}
  Q^\theta_t = e^{
      \eta \left(
      (\gamma_t - y_t)^2
        -
      (\xi^\theta_t - y_t)^2
    \right)}
  \end{equation*}
  is a forecast-continuous sequence having the defensive property.
\end{lemma}
\BP
  The continuity is obvious. We need to prove that
  \begin{equation}\label{eq:DFexpectsquare}
    pe^{\eta\left((\gamma - Y_2)^2 - (\xi_t^\theta - Y_2)^2\right)}
    +
    (1-p)e^{\eta\left((\gamma - Y_1)^2 - (\xi_t^\theta - Y_1)^2\right)}
    \le
    1
  \end{equation}
  holds for all~$\gamma\in[Y_1,Y_2]$ and $\eta\in\left(0,\frac{2}{(Y_2-Y_1)^2}\right]$.
  Indeed,
  for any $\gamma \in \R \setminus [Y_1,Y_2]$
  there exists $\tilde \gamma \in \{Y_1,Y_2\}$ such that
  $(\tilde \gamma - y)^2 \le (\gamma - y)^2$
  for any $y\in\Omega$.
  Since the exponent function is increasing,
  the inequality~\eqref{eq:DFexpectsquare} for any $\gamma\in\R$
  will follow.

  We use the correspondence~\eqref{eq:pigamma},
  $\xi_t^\theta = q(Y_2-Y_1) + Y_1$ for some $q \in \R$,
  and $\mu = \eta (Y_2-Y_1)^2$.
  Then we have to show that for all $p\in[0,1]$, 
  $q\in\R$ and $\eta\in\left(0,\frac{2}{(Y_2-Y_1)^2}\right]$
  \begin{equation*}
    pe^{\mu\left((1-p)^2 - (1-q)^2\right)}
    + (1-p)e^{\mu\left(p^2 - q^2\right)}
    \le 1.
  \end{equation*}
  If we substitute~$q=p+x$,
  the last inequality will reduce to
  \begin{equation*}
  pe^{2\mu(1-p)x}+(1-p)e^{-2\mu px}\le
   e^{\mu x^2},
    \quad
    \forall x\in\R.
  \end{equation*}
  Applying Hoeffding's inequality (see \cite{Hoeffding1963})
  to the random variable~$X$ that is
  equal to~$1$ with probability~$p$ and to~$0$ with probability~$(1-p)$,
  we obtain
  \begin{equation*}
  pe^{h(1-p)}+(1-p)e^{-hp}
  \le
  e^{h^2/8}
  \end{equation*} for any~$h\in\R$.
  With the substitution $h:=2\mu x$
  it reduces to
  \begin{equation*}
  pe^{2\mu(1-p)x}+(1-p)e^{-2\mu px}
  \le
  e^{\mu^2 x^2 / 2}
  \le
  e^{\mu x^2},
  \end{equation*}
  where the last inequality holds if $\mu \le 2$.
  The last inequality is equivalent to $\eta \le \frac{2}{(Y_2-Y_1)^2}$,
  which we assumed.
\EP
We will further use the maximum value for $\eta$, $\eta=\frac{2}{(Y_2-Y_1)^2}$.

The following lemma states
the most important for us property
of the sequences having the defensive property
originally proven in~\citet{Levin1976}.
\begin{lemma}\label{lem:Levin}
  Let $R$ be a forecast-continuous sequence having the defensive property.
  For any $T$
  there exists $p\in[0,1]$ such that
  for all $y \in \delta\Omega$
  \begin{equation*}
    R_T(\gamma^p,y)
    \le
    1.
  \end{equation*}
\end{lemma}
\BP
  Define a function $f_t: \delta\Omega \times [0,1] \to (-\infty,\infty]$ by
  \begin{equation*}
  f_t(p,y) =
    R_t(\gamma^p,y)
    -
    1.
  \end{equation*}
  Since $R$ is forecast-continuous
  and the correspondence~\eqref{eq:pigamma} is continuous,
  $f_t(y,p)$ is continuous in $p$.
  Since $R$ has the defensive property, we have
  \begin{equation}\label{eq:fexpect}
  p f(p,Y_2) + (1-p) f(1-p,Y_1) \le 0
  \end{equation}
  for all $p \in [0,1]$.
  In particular, $f(0,Y_1) \le 0$ and $f(1,Y_2) \le 0$.

  Our goal is to show that for some $p \in [0,1]$
  we have $f(p,Y_1) \le 0$ and $f(p,Y_2) \le 0$.
  If $f(0,Y_2) \le 0$,
  we  can take $p = 0$.
  If $f(1,Y_1) \le 0$,
  we can take $p = 1$.
  Assume that $f(0,Y_2) > 0$ and $f(1,Y_1) > 0$.
  Then the difference
  \begin{equation*}
  f(p) := f(p,Y_2) - f(p,Y_1)
  \end{equation*}
  is positive for $p = 0$ and negative for $p = 1$.
  By the intermediate value theorem,
  $f(p) = 0$ for some $p \in (0,1)$.
  By~\eqref{eq:fexpect} we have $f(p,Y_2) = f(p,Y_1) \le 0$.
\EP
This lemma shows that
at each step there is a probability measure
(corresponding to $p\in[0,1]$) such that
the sequence having the defensive property
remains less than one for any outcome.

The proof of the upper bounds for Defensive Forecasting
is based on the following argument.
\begin{lemma}\label{lem:Qdefprop}
  Assume that the sequence of
  functions~$Q^\theta_t$
  is forecast-continuous and has the defensive property.
  Then the mixtures $Q_t$ as functions of two variables $y,\gamma$ at the step $t$
  form a forecast-continuous sequence
  having the defensive property.
\end{lemma}
\BP
  The continuity easily follows from
  the continuity of $Q^\theta_t$ and the integration functional.
  We proceed by induction in $T$.
  For $T=0$ we have $\EXP_\pi Q_0 = \EXP_\pi 1 \le 1$.
  For $T>0$ assume that
  for any $y_1,\ldots,y_{T-2}\in\delta\Omega$
  and any $\gamma_1,\ldots,\gamma_{T-2}\in\Gamma$
  \begin{equation*}
  \EXP_\pi Q_{T-1} (y_1,\gamma_1,\ldots,y_{T-2},\gamma_{T-2},y,\gamma^{p^\pi}) \le 1
  \end{equation*}
  for any $\pi \in \PPP(\delta\Omega)$.
  Then by Lemma~\ref{lem:Levin}
  there exists $\pi_{T-1}\in \PPP(\delta\Omega)$ such that
  \begin{equation}\label{eq:QTm1less1}
  Q_{T-1}(y_1,\gamma_1,\ldots,y_{T-2},\gamma_{T-2},y,\gamma^{p^{\pi_{T-1}}})
  =
  \int_\Theta \prod_{t=1}^{T-2} \left(Q^\theta_t\right)^{\prod_{i=t}^{T-2}\alpha_i}
  Q^\theta_{T-1} P_0(d\theta) \le 1
  \end{equation}
  for any $y\in\delta\Omega$.
  We denote $\gamma_{T-1} = \gamma^{p^{\pi_{T-1}}}$ and fix any $y_{T-1}\in\Omega$.
  We obtain
  \begin{equation*}
  \begin{split}
  &\EXP_\pi Q_T(y_1,\gamma_1,\ldots,y_{T-1},\gamma_{T-1},y,\gamma^{p^\pi})
  \\ &=
  \EXP_\pi \int_\Theta \prod_{t=1}^{T-1}
  \left(Q^\theta_t(\gamma_t,y_t)\right)^{\prod_{i=t}^{T-1}\alpha_i}
  Q^\theta_T (\gamma^{p^\pi},y) P_0(d\theta)
  \\ &=
  \int_\Theta \prod_{t=1}^{T-1}
  \left(Q^\theta_t(\gamma_t,y_t)\right)^{\prod_{i=t}^{T-1}\alpha_i}
    \left(\EXP_\pi Q^\theta_T(\gamma^{p^\pi},y) \right) P_0(d\theta)
  \\ &\le
  \int_\Theta \prod_{t=1}^{T-1}
  \left(Q^\theta_t(\gamma_t,y_t)\right)^{\prod_{i=t}^{T-1}\alpha_i} P_0(d\theta)
  \\ &=
  \int_\Theta \left(\prod_{t=1}^{T-2}
  \left(Q^\theta_t\right)^{\prod_{i=t}^{T-2}\alpha_i}
    Q^\theta_{T-1}\right)^{\alpha_{T-1}} P_0(d\theta)
  \\ &\le
  \left( \int_\Theta \prod_{t=1}^{T-2} \left(Q^\theta_t\right)^{\prod_{i=t}^{T-2}\alpha_i}
  Q^\theta_{T-1} P_0(d\theta) \right)^{\alpha_{T-1}}
  \le
  1.
  \end{split}
  \end{equation*}
  The first inequality holds because
  $\EXP_\pi Q^\theta_T (\gamma^{p^\pi},y) \le 1$
  for any $\pi \in \PPP(\delta\Omega)$.
  The penultimate inequality holds due to the concavity
  of the function $x^\alpha$ with $x>0$, $\alpha\in[0,1]$.
  The last inequality holds due to~\eqref{eq:QTm1less1}.
  This completes the proof.
\EP

By Lemma~\ref{lem:Levin} at each step~$t$
there exists a prediction~$\gamma_t$ such that~$Q_t$
is less than one.
Now we only need to generalize Lemma~\ref{lem:Levin}
for the case when the outcome set is the full interval:
$\Omega = [Y_1,Y_2]$.
\begin{lemma}\label{lem:Levinfin}
  If $\gamma_T$ is such that
  $Q_T(y_1,\gamma_1,\ldots,y_{T-1},\gamma_{T-1},y,\gamma_T) \le 1$
  for all $y\in\{Y_1,Y_2\}$, then
  $Q_T(y_1,\gamma_1,\ldots,y_{T-1},\gamma_{T-1},y,\gamma_T) \le 1$
  for all $y\in[Y_1,Y_2]$.
\end{lemma}
\BP
  Note that any $y\in[Y_1,Y_2]$ can be represented as
  $y = uY_{T,2} + (1-u)Y_{T,1}$
  for some $u \in [0,1]$.
  Thus
  \begin{multline*}
  (\zeta_1 - y)^2 - (\zeta_2 - y)^2 = \zeta_1^2 - \zeta_2^2 - 2y(\zeta_1 - \zeta_2)\\
  = u[(\zeta_1 - Y_2)^2 - (\zeta_2 - Y_2)^2]
  + (1-u)[(\zeta_1 - Y_1)^2 - (\zeta_2 - Y_1)^2]
  \end{multline*}
  for any $\zeta_1,\zeta_2 \in \R$.
  Due to the
  convexity of the exponent function we have for any $\eta \ge 0$
  \begin{equation*}
  e^{\eta[(\zeta_1 - y)^2 - (\zeta_2 - y)^2]}
  \le ue^{\eta[(\zeta_1 - Y_2)^2 - (\zeta_2 - Y_2)^2]}
  + (1-u)e^{\eta[(\zeta_1 - Y_1)^2 - (\zeta_2 - Y_1)^2]}.
  \end{equation*}
  Thus
  \begin{equation*}
    Q^\theta_T(\gamma_T,y)
    \le
    uQ^\theta_T(\gamma_T,Y_2)
    +
    (1-u)Q^\theta_T(\gamma_T,Y_1)
  \end{equation*}
  and therefore
   \begin{multline*}
    Q_T(y_1,\gamma_1,\ldots,y_{T-1},\gamma_{T-1},y,\gamma_T)
    \le
    uQ_T(y_1,\gamma_1,\ldots,y_{T-1},\gamma_{T-1},Y_2,\gamma_T) \\
    +
    (1-u)Q_T(y_1,\gamma_1,\ldots,y_{T-1},\gamma_{T-1},Y_1,\gamma_T)
    \le 1
  \end{multline*}
  where the second inequality follows from the condition of the lemma.
\EP

Finally we obtain
\begin{equation}\label{eq:DFbound}
\int_\Theta \prod_{t=1}^{T-1}
e^{\eta \prod_{i=t}^{T-1} \alpha_i \left(\lambda(\gamma_t,y_t)
- \lambda(\xi_t^\theta,y_t)\right)}
 e^{\eta\left(\lambda(\gamma_T,y_T)
 - \lambda(\xi_T^\theta,y_T)\right)} P_0(d \theta) \le 1.
\end{equation}

\subsubsection{Derivation of the Prediction Strategies Using Defensive Forecasting}
Lemma~\ref{lem:Levin} describes an explicit strategy of making predictions.
This strategy relies on the search for a fixed point
and may become very inefficient especially for the cases of infinite number of experts.
Therefore we develop a more efficient strategies for each of our problems.

We first note that the strategy in Lemma ~\ref{lem:Levin} solves
\begin{multline*}
\int_\Theta \prod_{t=1}^{T-1} e^{\eta \prod_{i=t}^{T-1}\alpha_i \left( \lambda (\gamma_t,y_t)
-
\lambda(\xi_t^\theta,y_t)\right)} e^{\eta \left( \lambda(\gamma,Y_2)
-
\lambda(\xi_T^\theta,Y_2)\right)} P_0(d \theta)
\\ -
\int_\Theta \prod_{t=1}^{T-1} e^{\eta \prod_{i=t}^{T-1}\alpha_i \left( \lambda(\gamma_t,y_t)
-
\lambda(\xi_t^\theta,y_t)\right)} e^{\eta_T \left( \lambda(\gamma,Y_1)
-
\lambda(\xi_T^\theta,Y_1)\right)} P_0(d \theta)
= 0
\end{multline*}
in $\gamma\in[Y_1,Y_2]$ if the trivial predictions are not satisfactory
(the integral becomes a sum in the case of finite number of experts).
We define
\begin{equation}\label{eq:DFgenpred}
g_T(y) := -\frac{1}{\eta} \ln \int_{\Theta} e^{-\eta \lambda (\xi_T^\theta,y)}
\prod_{t=1}^{T-1} e^{-\eta \prod_{i=t}^{T-1}\alpha_i \lambda(\xi_t^\theta,y_t)} P_0(d \theta)
\end{equation}
for any $y\in\Omega$.
Rewriting the equation for the root we have
\begin{equation*}
e^{\eta \left( \lambda_T(\gamma,Y_2)
-
g_T(Y_2)\right) }
-
e^{\eta \left( \lambda_T(\gamma,Y_1)
-
g_T(Y_1)\right)}
=
0
\end{equation*}
Moving the second exponent to the right-hand side
and taking $\log_{\eta}$ of both sides we obtain
\begin{equation}\label{eq:DFeffpredeq}
\lambda(\gamma,Y_2) - g_T(Y_2)
=
\lambda(\gamma,Y_1) - g_T(Y_1).
\end{equation}

For the square loss we can solve \eqref{eq:DFeffpredeq} in $\gamma$:
\begin{equation}\label{eq:finpred}
\gamma = \frac{Y_2+Y_1}{2} - \frac{g(Y_2)-g(Y_1)}{2(Y_2-Y_1)}.
\end{equation}
This formula for predictions
is equivalent to~\eqref{eq:substsquare}.

\fi


\end{document}